\documentclass[10pt,twocolumn,letterpaper]{article}

\usepackage{3dv}
\usepackage{times}
\usepackage{epsfig}
\usepackage{graphicx}
\usepackage{amsmath,amssymb} 
\usepackage{color}

\usepackage{algorithm,algorithmicx,algpseudocode}
\usepackage{bm,xspace}
\usepackage{comment}
\usepackage{verbatim}
\usepackage{balance}
\usepackage{etoolbox,siunitx}
\usepackage{calc}
\usepackage{pifont,hologo}
\usepackage{adjustbox}
\usepackage{tabularx}
\usepackage{multirow}
\usepackage{subcaption} 
\usepackage{overpic}
\usepackage{array,makecell, booktabs}%
\usepackage[font=small]{caption}
\usepackage{hhline}
\usepackage{authblk} 

\def\eg{\emph{e.g.~}}

\def\etal{\emph{et al.~}}

\newenvironment{packed_item}{
\begin{itemize}
\vspace{0pt}
  \setlength{\itemsep}{0pt}
  \setlength{\parskip}{0pt}
  \setlength{\parsep}{0pt}
  \setlength{\topsep}{-10pt}
  \setlength{\partopsep}{0pt}
}{\end{itemize}}

\def\family{MS}

\usepackage[pagebackref=true,breaklinks=true,letterpaper=true,colorlinks,bookmarks=false]{hyperref}

\threedvfinalcopy 

\def\threedvPaperID{36} 

\ifthreedvfinal\fi
\begin{document}

\title{Matching-space Stereo Networks for Cross-domain Generalization}

\author[*]{Changjiang Cai} 
\author[$\dagger$]{Matteo Poggi}
\author[$\dagger$]{Stefano Mattoccia}
\author[*]{Philippos Mordohai}


{ 
\makeatletter
\renewcommand\AB@affilsepx{\qquad  \protect \Affilfont}
\makeatother
\affil[*]{\protect Stevens Institute of Technology} 
\affil[$\dagger$]{University of Bologna \protect \\}
\affil[ ]{ {\tt\small $^*$ccai1@stevens.edu, $^*$mordohai@cs.stevens.edu, \tt\small $^\dagger$\{m.poggi,stefano.mattoccia\}@unibo.it }}
}

\renewcommand\Authands{ \qquad } 
\renewcommand{\Authsep}{\qquad} 

\maketitle


\begin{abstract}
End-to-end deep networks represent the state of the art for stereo matching. While excelling on images framing environments similar to the training set, major drops in accuracy occur in unseen domains (e.g., when moving from synthetic to real scenes). In this paper we introduce a novel family of architectures, namely Matching-Space Networks (MS-Nets), with improved generalization properties. By replacing learning-based feature extraction from image RGB values with matching functions and confidence measures from conventional wisdom, we move the learning process from the color space to the Matching Space, avoiding over-specialization to domain specific features. Extensive experimental results on four real datasets highlight that our proposal leads to superior generalization to unseen environments over conventional deep architectures, keeping accuracy on the source domain almost unaltered.  Our code is available at \url{https://github.com/ccj5351/MS-Nets}.
\end{abstract}

\section{Introduction}\label{sec:intro}


The rising availability of stereo imagery with ground truth depth \cite{Geiger2012CVPR,menze2015object,scharstein2014high,schops2017multi} has enabled the development of machine learning based stereo matching algorithms. The first attempts to exploit machine learning for dense correspondence focused on matching \cite{batsos2018cbmv,chen2015deep,luo2016efficient,zbontar2016stereo} or other stages \cite{batsos2018recresnet,gidaris2017DRR,park2015leveraging,poggi2016learning,seki2017sgm-net} of the stereo pipeline \cite{scharstein2002taxonomy} and were combined with conventional components.
As in other areas of computer vision, end-to-end methods \cite{mayer2016large} soon became the dominant paradigm. 
They can be distinguished in two categories according to network architecture: 2D \cite{liang2018learning_1st_Rob,mayer2016large,pang2017cascade,song2019edgestereo,yang2018segstereo} and 3D \cite{chang2018psmnet,kendall2017-gcnet,yu2018deep,zhang2019ga} convolutional networks. The latter reason about geometry by building a \textit{matching volume}, either correlating or concatenating \textit{learned features} from the images. 3D convolutions enable these methods to consider context beyond a fixed disparity value (a slice of the matching volume) for higher accuracy.

\begin{figure*}[t]
    \centering
    \renewcommand{\tabcolsep}{1pt}
    \scriptsize
    \begin{tabular}{cccc}
        \includegraphics[width=0.25\textwidth]{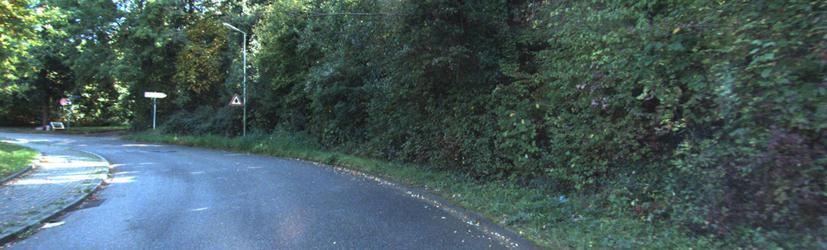} &
        \includegraphics[width=0.25\textwidth]{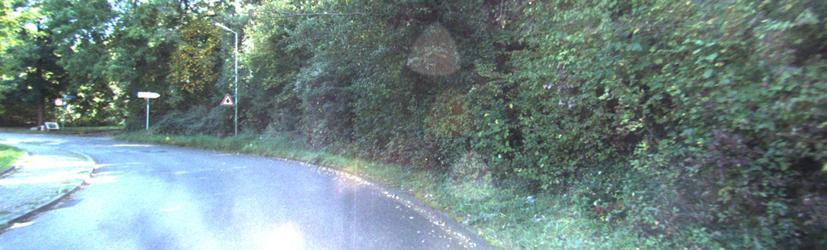} &
        \includegraphics[width=0.25\textwidth]{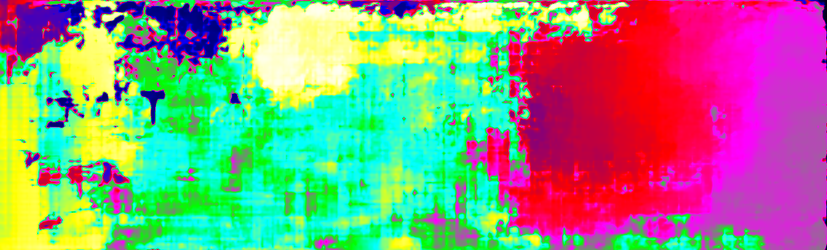} &
        \includegraphics[width=0.25\textwidth]{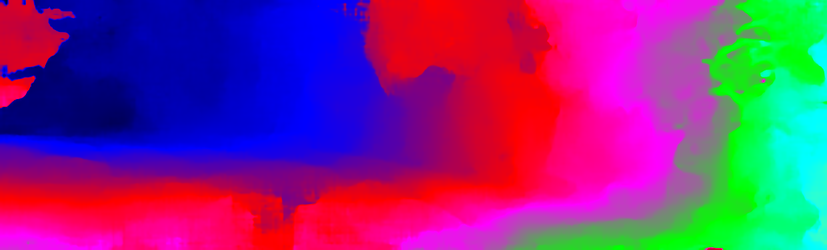} 
        \\
        (a) & (b) & (c) & (d)\\
    \end{tabular}
    \vspace{-8pt}
    \caption{\textbf{Generalization across domains.} (a) and (b) Challenging stereo pair from the KITTI dataset. (c) Disparity map estimated by PSMNet \cite{chang2018psmnet}. (d) Disparity map estimated by our \family-PSMNet. Both networks are trained on the same \textbf{synthetic data.}}
    \label{fig:teaser}
      \vspace{-8pt}
\end{figure*}

Although end-to-end models excel at specializing on specific domains when enough data are available for training, \eg autonomous driving \cite{Geiger2012CVPR,menze2015object}, they are less effective at generalization to very different domains or with high variety of image content. Strong evidence supporting this emerges by browsing online benchmarks: while on KITTI 2012 \cite{Geiger2012CVPR} and 2015 \cite{menze2015object} end-to-end networks dominate the leaderboards, very few of them appear on the Middlebury 2014 leaderboard \cite{scharstein2014high} and typically achieve lower accuracy than hand-designed algorithms \cite{taniai2018local-expansion} deploying machine learning on individual steps \cite{zbontar2016stereo} of the pipeline. This is due to the diversity of the Middlebury dataset.



Poor generalization is a cause for concern, since gathering annotated data may be too expensive or infeasible in practical applications. 
We argue that this lack of generalization, or \textbf{over-specialization}, is caused by the learning process being driven by image content, i.e. the network learns how to match pixels by strongly relying on appearance properties. When such content differs substantially from the one observed in training, end-to-end approaches suffer large accuracy drops. Usually large amounts of labelled images, often generated in synthetic environments \cite{mayer2016large}, are used to improve generalization and mitigate this effect. On the other hand, domain shifts still pose difficulties \cite{pang2018zoom-and-learn,tonioni2017unsupervised,Tonioni_2019_CVPR}, in particular when moving from synthetic to real imagery affected by reflective surfaces, sensor noise and illumination conditions, which have not been modeled in the simulators.
Anyway, there is evidence in literature that this effect can be soften by moving the learning process to different representations \cite{batsos2018cbmv} or parts of the pipeline \cite{schonberger2018sgm-forest}.
We claim that better generalization can be achieved by choosing a representation insensitive to common variations of the input images. Instead of augmenting the dataset to guide the network to achieve certain invariances, one could design a hybrid approach in which some invariances are learned from the data while others are imposed by the design. For example, if we wish the network to be invariant to affine transformations of image intensity or color, we would use normalized cross correlation (NCC) in a conventional stereo algorithm. In an end-to-end trainable algorithm, we could either achieve this invariance via augmenting the data by affinely transforming one of the images, or by using an NCC-like abstraction in the representation of the images.


In this paper, we propose a new family of end-to-end architectures designed to have invariant properties that make them robust to domain shifts.
Our networks learn to reason about stereo matching in the domain of matching functions by combining four matching functions and associated confidence scores \cite{hu2012quantitative,poggi2017quantitative}. Stacking multiple of these cues results in a 4D tensor compatible with matching volumes processed by 3D networks. This way, the network is never exposed to image appearance and is forced to learn in the \emph{Matching Space} (\family) only, avoiding over-specialization.

We demonstrate the effectiveness of the proposed representation by implementing two 3D convolutional architectures based on the above principle. Specifically, we replicate two popular, but different, 3D convolutional networks, GCNet \cite{kendall2017-gcnet} and PSMNet \cite{chang2018psmnet}, replacing their matching volume, which is based on deep image features, with the proposed \family{} representation. Extensive experiments show that our \family-Nets generalize better to data from domains that differ substantially from the training one. We believe that the reason for this is that \emph{our approach learns to reason on relationships in the matching volume without being affected by image appearance,} which our networks never observe directly.
Figure \ref{fig:teaser} shows disparity estimation by both PSMNet and its \family{} variant on a challenging stereo pair from KITTI, after being trained on synthetic images only. In particular, we can notice large changes of illumination between the two frames, possibly never observed during training. Estimated disparity maps show that such perturbations drive PSMNet to totally unreliable estimations. Conversely, \family-PSMNet learned a robust representation leading to accurate results.


Our claim is supported by extensive experimental results on five popular datasets: SceneFlow \cite{mayer2016large}, KITTI 2012 \cite{Geiger2012CVPR}, KITTI 2015 \cite{menze2015object}, Middlebury 2014 \cite{scharstein2014high} and ETH3D \cite{schops2017multi}. In particular, we study a variety of training protocols involving both large and limited amounts of labelled images in order to dig deeper into the relationship between training data and model performance, an aspect that has been largely ignored in previous works \cite{chang2018psmnet,kendall2017-gcnet,liang2018learning_1st_Rob,mayer2016large,song2019edgestereo,yang2018segstereo}. Our \family-Nets are able to generalize much better than GCNet and PSMNet with minor loss of accuracy in the source domain.
The main contributions of this paper are: 
\vspace{-6pt}
\begin{packed_item}
\item The observation and testing of the hypothesis that not exposing CNNs directly to image appearance leads to better generalization properties;
\item A novel family of architectures, \family-Nets, and one of its possible implementations built on conventional wisdom \cite{batsos2018cbmv} and popular 3D networks \cite{kendall2017-gcnet,chang2018psmnet};  
\item An extensive set of experiments highlighting the behavior of both 3D and \family-Nets under domain shift.
\end{packed_item}



\section{Related Work}\label{sec:related}
The problem of disparity estimation from stereo image pairs has been studied for decades. We refer readers to survey papers \cite{Geiger2012CVPR,janai2017computer,poggi2017quantitative,scharstein2002taxonomy}. In this section, we review deep learning based stereo methods, broadly divided into \textit{matching cost learning and optimization} and \textit{end-to-end dense disparity regression}. 

\noindent \textbf{Matching cost learning and optimization.} \, Convolutional Neural Networks for predicting the matching costs were first introduced by 
\cite{zbontar2016stereo}. The outputs of the MC-CNN network were refined according to the non-learned traditional pipeline \cite{mei2011building,scharstein2002taxonomy} to generate disparity maps. \cite{luo2016efficient} substituted an inner product layer for the fully connected layer in MC-CNN \cite{zbontar2016stereo} to alleviate the expensive computational burden. 
\cite{shaked2017improved} presented a network for matching cost computation, utilizing a highway network with multi-level weighted residual shortcuts. A deep embedding model presented by \cite{chen2015deep} was learned from a multi-scale ensemble framework, which fuses feature vectors learned at different scales. \cite{seki2017sgm-net} used CNNs to learn the penalty-parameters of the Semi-Global Matching (SGM) algorithm \cite{hirschmuller08} and proposed a new parameterization of the same algorithm discriminating between positive and negative disparity changes. 
In contrast, our method, combining conventional matching functions and confidence measures as matching cost, can be trained end-to-end. 


\noindent \textbf{End-to-end disparity regression.} \, Mayer \etal \cite{mayer2016large} were the first to propose an end-to-end stereo network, namely DispNet, with an encoder-decoder architecture for disparity estimation trained on a large synthetic dataset created by them. This synthetic dataset enabled the development of end-to-end deep stereo networks. \cite{kendall2017-gcnet} presented GCNet to exploit contextual information for disparity regression, via 3D-convolution on the matching volume using deep unary features and a differentiable \textit{soft-argmin} operation. Chang and Chen put forward PSMNet \cite{chang2018psmnet} consisting of spatial pyramid pooling for unary feature extraction, and stacked 3D hourglasses for matching volume regularization. StereoDRNet \cite{Chabra_2019_CVPR} extends PSMNet \cite{chang2018psmnet} by replacing the spatial pyramid pooling \cite{he2015spatial} with vortex pooling \cite{xie2018vortex} in feature extraction, and by utilizing 3D dilated convolutions in cost volume filtering. 
\cite{yu2018deep} learned cost aggregation via a two-stream network for generation and selection of cost aggregation proposals. In parallel, more architectures built on DispNet have been proposed, leveraging cascade residual learning \cite{pang2017cascade} or deploying multi-task learning by jointly learning disparity estimation together with edge detection \cite{song2019edgestereo} or semantic segmentation \cite{yang2018segstereo}. \cite{liang2018learning_1st_Rob}
incorporated matching cost calculation and aggregation, disparity estimation and refinement into one network, ranking first at the Robust Vision Challenge 2018\footnote{http://www.robustvision.net/leaderboard.php?benchmark=stereo}.  \cite{guo2019group} constructed the matching volume by group-wise correlation of features which were split into multiple groups along the channel dimension. Zhang \etal \cite{zhang2019ga} proposed GA-Net, a deep guided aggregation network, including semi-global aggregation (SGA) and local guided aggregation (LGA) layers for efficient end-to-end stereo matching. Our proposed networks, never exposed to RGB values, are trained end-to-end from the 
matching space to final disparity maps. 

\noindent \textbf{Deep learning for domain transfer.} \, Most deep stereo models are particularly data dependent and their performance drops considerably when dealing with unseen domains different from those observed during training \cite{tonioni2017unsupervised,tonioni2020unsupervised}. To tackle the domain shift problem, two main strategies are involved: image synthesis \cite{chen2017photographic,hoffman2018cycada,liang2019unsupervised}, and un-/self-supervised adaptation \cite{casser2019struct2depth,guo2018learning,pang2018zoom-and-learn,tonioni2017unsupervised,tonioni2020unsupervised,Tonioni_2019_CVPR_learn_adapt,Tonioni_2019_CVPR,zhong2017self,zhong2018open}. In contrast, our method aims at being transferred \textbf{without adaptation} to different domains, being this possibility more appealing for practical applications.


\begin{figure*}
\begin{center}
\includegraphics[width=0.8\linewidth]{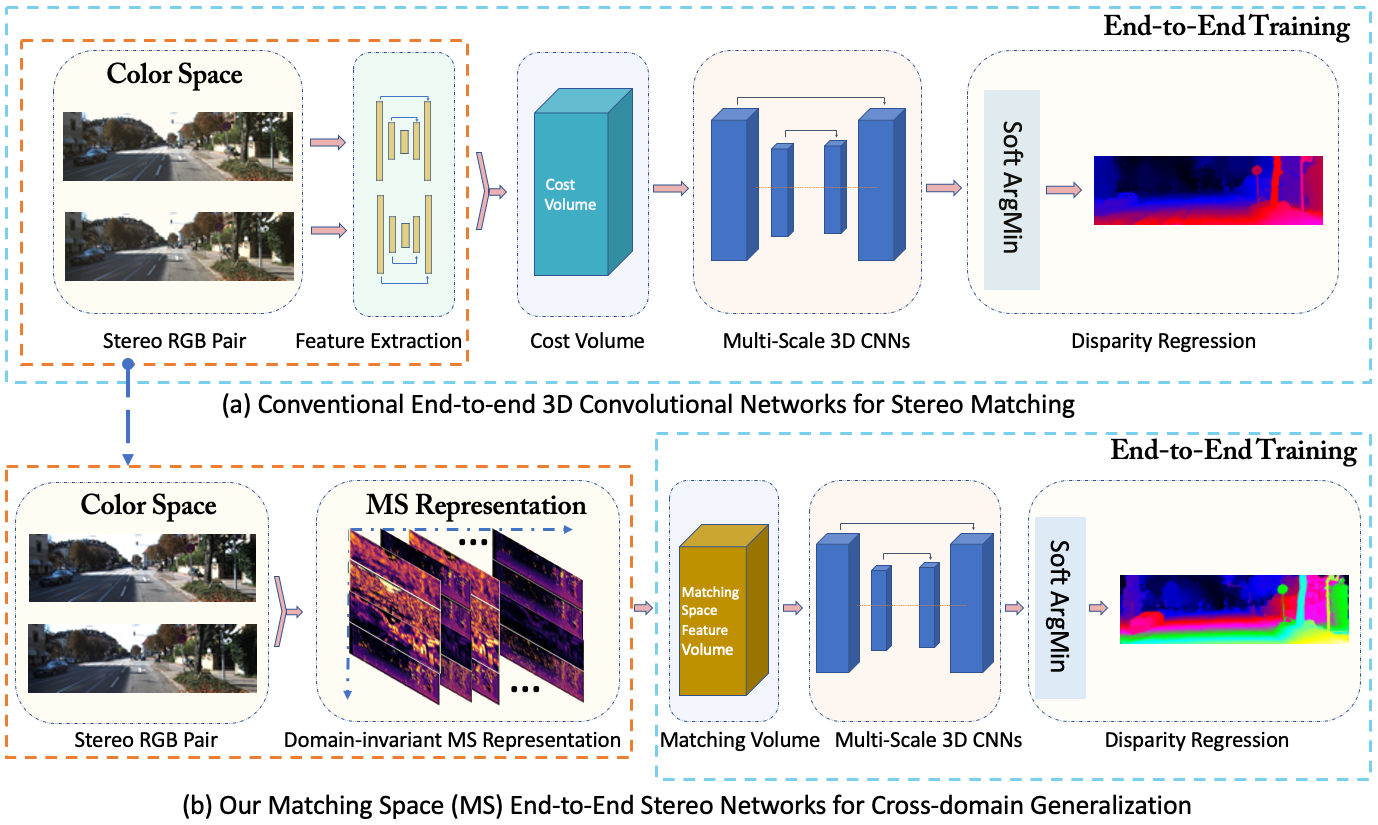}
\end{center}
\vspace{-18pt}
\caption{\textbf{Architecture overview.} (a) shows a conventional end-to-end 3D convolutional network for stereo matching. (b) illustrates our end-to-end \family-Nets architecture. 
The orange dashed boxes show the different feature extractors of the two architectures. The blue dashed boxes show the part of each network that is trained end-to-end. Feature extractors in \family-Nets are fixed and isolate the network from RGB.
}
\label{fig:cbmvnet-archi}
  \vspace{-8pt}
\end{figure*}

\section{Approach}\label{sec:approach}
{\indent Deep learning methods \cite{chang2018psmnet,kendall2017-gcnet,mayer2016large,zbontar2016stereo,zhong2017self} work extremely well on disparity estimation when sufficient data with ground truth are available. However, they have been proven vulnerable to out-of-distribution data. When dealing with unseen environments quite different from those employed to train the network, the accuracy may rapidly decrease.} 
By moving the learning process to Matching Space, we force the neural network to learn a more general representation, avoiding over-fitting to specific image appearance statistics that are characteristic of the training domain and thus to be more robust to such decreases.

For one possible implementation of \family-Nets, we select a subset of the CBMV features \cite{batsos2018cbmv}, which have shown good \textbf{generalization} across distinct datasets with respect to neural networks \cite{zbontar2016stereo} when used to learn a matching function.
In contrast to existing end-to-end networks, which are directly exposed to raw image intensities (or RGB values) \cite{chang2018psmnet,kendall2017-gcnet,mayer2016large,zbontar2016stereo,zhong2017self}, the proposed \family-Nets, leveraging features that encode geometric constraints and prior knowledge, can be transferred without adaptation to different domains. The general architecture of the proposed family of networks is shown in Fig. \ref{fig:cbmvnet-archi}.

As base architectures, we have chosen PSMNet \cite{chang2018psmnet} and GCNet \cite{kendall2017-gcnet}. Both belong to the 3D CNN category, but they have different architectures and parameter configuration. Specifically, PSMNet has 3.3M ($63.5\%$) parameters in the unary feature extraction modules, versus only 1.9M ($36.5\%$) in the 3D CNN layers for cost volume regularization. On the other hand, GCNet has the opposite configuration, i.e. 0.3M ($11.5\%$) parameters for feature extraction and 2.3M ($88.5\%$) parameters in the 3D CNNs. We will show how these differences impact generalization.

\subsection{Matching-Space Features and Volume}
We follow the notation of \cite{batsos2018cbmv} in this section.
Given a rectified stereo pair comprising the left and right image, $I_L$ and $I_R$, a \textit{ matching hypothesis} $(x_L, x_R, y)$ represents a potential correspondence between a pixel $p_L(x_L, y)$ in $I_L$ and a pixel $p_R(x_R = x_L -d, y)$ in $I_R$, with disparity $d$ defined as $d = x_L - x_R$. For each disparity, we adopt eight features, consisting of the raw matching cost and likelihood for four matchers. The matchers used are: the normalized cross correlation (\textit{NCC}), the zero-mean sum of absolute differences (\textit{ZSAD}), the census transform (\textit{CENSUS}) and the absolute differences of the horizontal Sobel operator (\textit{SOBEL}). 
Unlike \cite{batsos2018cbmv}, we only use the left-to-right-likelihood for each matcher to keep memory and processing requirements manageable. 

According to \cite{hu2012quantitative} and \cite{batsos2018cbmv}, the likelihood, a confidence measure of all disparities of a given pixel $p_L$, is obtained by converting its cost curve to a probability density function for disparity $d$ under consideration. Using \textit{ZSAD} ($z$ in short in Eq. \ref{eq:likelihood}) as an example, with the left image $I_L$ as reference and the right image $I_R$ as the matching target, the likelihood is defined as:

\vspace{-0.8cm}
\begin{equation}\label{eq:likelihood}
L_{z}(x_L,y,d)=\frac{\exp{\big( -\frac{(C_{z}(x_L, y, d)-C_{z,min})^2}{2 \sigma_{z}^2}} \big) }{\sum_{i} \exp{ \big(  -\frac{ (C_{z}(x_L,y, d_i) - C_{z,min})^2}{2 \sigma_{z}^2} \big) }}
\end{equation}

\noindent where $C_{z,min}$ denotes the minimum cost of \textit{ZSAD} for the left pixel $p_L$, and $\sigma_{_{z}}$ is a predefined hyper-parameter that depends on the corresponding matching algorithm.

\noindent \textbf{Matching Volume.} \,\, Given the above  features extracted from a stereo image pair at size $H \times W $, we generate a 4D matching volume of dimensionality $D \times H \times W \times F$, across each disparity level, where $F$ is the number of features  (i.e., F = 8 in this case). 
The matching values for  \textit{CENSUS}, \textit{ZSAD}, \textit{SOBEL} and \textit{NCC} are normalized to $[0, 1]$ before being fed into the subsequent multi-scale 3D CNN encoder-decoder layers. 

\subsection{Multi-Scale 3D CNN Encoder-Decoder}


The generated 4D matching volume is too noisy to directly predict disparity maps in Winner-Take-All fashion, due to textureless or reflective regions among other challenges. Hence, the matching volume is regularized via 3D multi-scale encoder-decoder architectures \cite{chang2018psmnet,kendall2017-gcnet} for effective disparity optimization. %
3D networks \cite{chang2018psmnet,kendall2017-gcnet} perform this after such a volume has been derived from earlier subnetworks exposed to image appearance, thus prone to over-fitting and poor generalization. In our case, the volume itself is the input to the first layers of the networks, that starts the entire learning process in the Matching Space and thus prevents exposing the networks to image appearance. We present two variants of 3D networks for matching volume regularization based on GCNet and PSMNet.


\textbf{\family-GCNet variant.} \, 
We adopt the regularization sub-network of GCNet \cite{kendall2017-gcnet}, and feed the \family{} matching volume to it for end-to-end disparity estimation.
Specifically, given a stereo pair of size $H \times W$ and disparity range $D$, we first extract the features from the image pair, after down-sampling by a factor of 2, to compute a matching volume of size $D/2 \times H/2 \times W/2 \times F$. 
Then the generated matching volume is regularized through a 4-level down-sampling (via 3D convolution with stride 2) in the encoder, and a corresponding 4-level up-sampling (via 3D \textit{transposed} convolution with stride 2) in the decoder. The 4-level down-sampling, together with the input images, down-sampled by a factor of 2 (for feature extraction), results in a total of 32-times enlarged receptive field in order to exploit context information. 
In our implementation the matching volume is encoded into a $D/32 \times H/32 \times W/32 \times 16F$ volume, and then decoded to $D/2 \times H/2 \times W/2 \times 4F$. In order to make the best use of ground-truth disparity maps at the original resolution, we apply another 3-D transposed convolution (with stride 2) and a single feature (i.e., the channel dimension) output resulting in the final regularized $D \times H \times W$ matching volume, essential for dense disparity estimation in the original input dimensions. 

\textbf{\family-PSMNet variant.} \,\, We also implement a variant of PSMNet \cite{chang2018psmnet} by replacing the spatial pyramid pooling layers in charge of extracting deep features directly from RGB images with matching volumes from the Matching Space.

The initial $D/2 \times H/2 \times W/2 \times F$ 4D volume, obtained as for \family-GCNet, is down-sampled to quarter resolution by means of two $3\times3\times3$ convolution layers, the first with stride 2, in order to reduce the computational burden, then is processed by two more $3\times3\times3$ layers extracting 32 features each. Then, following \cite{chang2018psmnet}, we regularize the 4D volume through a stacked-hourglass architecture, built of three encoding-decoding blocks made of four $3\times3\times3$ convolution layers, with strides respectively 2, 1, 2 and 1, and two $3\times3\times3$ transposed convolution layers restoring the input resolution. Each hourglass generates a regularized volume, from which an intermediate disparity is obtained, and is implemented exactly as in the original paper \cite{chang2018psmnet} (which we refer the reader to for the sake of space) to ensure a fair comparison. While at training time the loss function is computed on all three intermediate results, at test time only the one obtained from the last hourglass is used as output.


\subsection{Disparity Regression}
In both cases, we use the differentiable \textit{soft argmin} as proposed by \cite{kendall2017-gcnet}
to regress the disparity from the regularized matching volume. This enables \textit{end-to-end} training and \textit{continuous} disparity maps as output. The cost curve $C_d$ for a given pixel $p_L$ is first converted to a probability of each disparity $d \in [0, D]$, via the \textit{softmax} operation $\sigma (\cdot)$. Then the predicted disparity $\hat{d}$ is calculated as the \textit{expected} value (i.e., the probability-weighted average) of random variable $d \in [0, D]$, defined as

\begin{equation}\label{eq:softargmin}
\hat{d} = \sum_{d=0}^{D} d \times \sigma(-C_d)
\end{equation}

\vspace{-8pt}

\subsection{Loss Function}
Our network is trained end-to-end, via supervised learning using datasets with 
ground truth disparities. The loss is evaluated and averaged only over the valid pixels (i.e., with ground truth disparity). For the \family-GCNet variant, we follow the authors of GCNet and use the $L_1$ loss, defined as: 

\begin{equation}\label{eq:l1_loss}
L(d, \hat{d}) = \frac{1}{N} \sum_{i=1}^N || d_i - \hat{d_i} ||_1 
\end{equation}

\noindent where $N$ is the number of valid pixels. 

For \family-PSMNet, we adopt the smooth $L_1$ loss, as in PSMNet. 

\vspace{-0.6cm}
\begin{align}\label{eq:l1_loss_smooth_a}
L_s(d, \hat{d}) &= \frac{1}{N} \sum_{i=1}^N  \text{smooth}_{L_1}(d_i - \hat{d_i}) \nonumber \\
\text{smooth}_{L_1}(x) &= 
\begin{cases}
0.5x^2 & \text{if } |x| < 1 \\
|x| - 0.5 & \text{otherwise}
\end{cases}
\end{align}

\noindent with $s = 0, 1, 2$ for the three intermediate outputs.
The total loss 
is defined as the weighted sum of the three intermediate losses 
with weights $0.5$, $0.7$, and $1.0$, respectively.

\section{Experimental Results}\label{sec:experiments}

In this section, we evaluate our algorithms in a domain transfer setting on five datasets: 
Scene Flow (\textit{SF}) \cite{mayer2016large}, KITTI 2012 (\textit{KT12}) \cite{Geiger2012CVPR}, KITTI 2015 (\textit{KT15}) \cite{menze2015object}, Middlebury 2014 (\textit{MB}) \cite{scharstein2014high}, and the ETH3D stereo benchmark (\textit{ETH3D}) \cite{schops2017multi}. From now on, we use \textit{domain} and \textit{dataset} interchangeably. For convenience, we adopt the notation $\mathcal{S} \rightarrow \mathcal{T}$ to describe that the model, trained in source domain $\mathcal{S}$, is transferred \textbf{without adaptation} to a different target domain $\mathcal{T}$. 

\begin{table*}[t]
\small
\centering
\centering
\scalebox{0.85}{
	\begin{tabularx}{\linewidth}{lXXXXXXXXX}
	\hline
	&	&	\multicolumn{8}{c}{Target domain $\mathcal{T}$} \\
	& & \multicolumn{2}{c}{KT12 (bad3-noc)\%} & \multicolumn{2}{c}{KT15 (bad3-all)\%} & \multicolumn{2}{c}{MB (bad2-noc)\%} & \multicolumn{2}{c}{ETH3D (bad1-noc)\%} \\
				\hline
	\multirow{5}{*}{\rotatebox{90}{Source domain $\mathcal{S}$ \,\,\,}}
	& & GCNet& \family-GCNet & GCNet & \family-GCNet & GCNet & \family-GCNet & GCNet & \family-GCNet \\ 
	\cline{2-10} 

	& \multicolumn{1}{r}{sf-all}  & 6.22 &\textbf{5.51}  &14.68 &\textbf{6.21} &30.42 & \textbf{18.52}  &\textbf{8.03} & 8.84 \\
	
	& \multicolumn{1}{r}{sf-3k} &7.40 &\textbf{6.51}  &17.43 &\textbf{7.77} &34.73 &\textbf{21.82}  &15.57 &\textbf{14.81} \\

	& \multicolumn{1}{r}{sfD3k} &\textbf{8.50}  & 10.50  &13.68 &\textbf{12.15} &42.56 &\textbf{25.59}  &\textbf{17.64} & 19.29 \\
	
	& \multicolumn{1}{r}{sfM3k} &8.29 &\textbf{7.98}  & 9.51 &\textbf{8.79} &28.05 &\textbf{24.21} &\textbf{11.17} &14.4 \\
	
	& \multicolumn{1}{r}{sfF3k} &8.48 &\textbf{7.78} & 37.32 &\textbf{9.11} &41.30 &\textbf{20.77} &20.87 &\textbf{17.41} \\
	\hline
	
\end{tabularx}
}
\vspace{-4pt}
\caption*{ (a) Generalization results for GCNet and \family-GCNet.}
\label{tab:scarce_labeled_data_gcnet}

\centering
\scalebox{0.85}{
	\begin{tabularx}{\linewidth}{lXXXXXXXXX}
	\hline
	&	&	\multicolumn{8}{c}{Target domain $\mathcal{T}$} \\
	& & \multicolumn{2}{c}{KT12 (bad3-noc)\%} & \multicolumn{2}{c}{KT15 (bad3-all)\%} & \multicolumn{2}{c}{MB (bad2-noc)\%} & \multicolumn{2}{c}{ETH3D (bad1-noc)\%} \\
				\hline
    \multirow{5}{*}{\rotatebox{90}{Source domain $\mathcal{S}$ \,\,\,}}
	& & PSMNet& \family-PSMNet & PSMNet & \family-PSMNet & PSMNet & \family-PSMNet & PSMNet & \family-PSMNet \\

	\cline{2-10} 
    & \multicolumn{1}{r}{sf-all}  &27.02 &\textbf{13.97} &26.62 
    & \textbf{7.76}  & 26.92 
    &\textbf{19.81}  &18.91 &\textbf{16.84} \\
	& \multicolumn{1}{r}{sf-3k} &35.63 &\textbf{8.57}  &35.56 
	&\textbf{8.36} &32.48 &\textbf{19.44} 
	&19.44 &\textbf{15.36}  \\
	& \multicolumn{1}{r}{sfD3k} &40.12 &\textbf{17.81} &39.00  
	&\textbf{16.39} &37.14 
	&\textbf{24.78}  &\textbf{19.58} &22.29 \\
	& \multicolumn{1}{r}{sfM3k} & 7.93 &\textbf{7.68}  & 8.20  
	&\textbf{7.00} &24.70  
	&\textbf{20.49} &14.58 &\textbf{14.24} \\
	& \multicolumn{1}{r}{sfF3k} & 45.41 &\textbf{9.14}  & 49.50  
	&\textbf{8.52} & 33.33 
	&\textbf{20.39} & 30.14 &\textbf{16.96} \\
	\hline
\end{tabularx}
}
\vspace{-5pt}
\caption*{(b) Generalization results for PSMNet and \family-PSMNet.}
  \label{tab:scarce_labeled_data_psmnet}
  \vspace{-6pt}
\caption{\textbf{Generalization results for \family-Nets, GCNet and PSM-Net.} The bad-$x$ errors are evaluated on $\mathcal {T}$ after training on $\mathcal {S}$ without any fine-tuning or adaptation to $\mathcal {T}$, using the default error measure of $\mathcal{T}$.} 
\label{tab:scarce_labeled_data}
\vspace{-6pt}
\end{table*}

\begin{packed_item}
     \item \textbf{Scene Flow} is a large synthetic dataset which contains 3 subsets - Driving (\textit{sfD}), Monkaa (\textit{sfM}), and FlyingThings3D (\textit{sfF}) 
     totaling more than 39000 stereo frames with dense ground truth disparity maps (35454 for training, and 4370 for testing) at $960 \times 540$ pixel resolution. In our experiments, Scene Flow is used to train networks from scratch.
    \item \textbf{KITTI} is a real-world dataset with two versions: KT12 (including 194 training and 195 testing stereo pairs) and KT15 (including 200 training and 200 testing stereo pairs), both approximately at $1240 \times 376$ pixel resolution. Compared to KT12, KT15 provides more dense ground truth disparity for cars and windshields. We use the training sets to evaluate all networks, but not for training. 
    \item \textbf{Middlebury 2014} consists of 15 training and 15 testing stereo pairs, as well as 12 additional stereo pairs with available ground truth. The ground truth for the test set is withheld. We use the half-resolution images of the training set to evaluate all networks.
    \item \textbf{ETH3D Low-res two-view} provides 27 training and 20 testing stereo pairs.
    As above, we use the training set as test data. 
\end{packed_item}

GCNet \cite{kendall2017-gcnet} and PSMNet \cite{chang2018psmnet} are the baselines. We implement GCNet and  \family-GCNet using PyTorch. The correctness of our GCNet implementation has been verified via a D1-all (i.e., percentage of outliers in all pixels of the reference frame with ground truth)
error of 2.76\% (versus 2.87\% as reported by \cite{kendall2017-gcnet}) on the KT15 benchmark. The PyTorch source code of PSMNet
is provided by the authors \cite{chang2018psmnet}. Therefore, we implemented \family-PSMNet using their code as the starting point. For fair comparison with the baselines, we adopt their training schemes. Specifically, \family-GCNet is trained end-to-end with RMSProp and a constant learning rate of $1 \times 10^{-3}$, while \family-PSMNet is trained using Adam ($\beta_1 = 0.9, \beta_2 = 0.999$), with a constant learning rate of $1 \times 10^{-3}$.

In both cases, we configure the hyper-parameters of the CBMV features following \cite{batsos2018cbmv}, i.e., $\sigma_{NCC}=0.1$, $\sigma_{ZSAD}=100$, $\sigma_{CENSUS}=8$ and $\sigma_{SOBEL}=100$. The matching windows are $3\times3$, $5\times5$, $11\times11$ and $5\times5$ for NCC, ZSAD, CENSUS and SOBEL, respectively. (Sensitivity to the sizes of the matching windows is low and varying these parameters is out of the scope of this paper. They remain unchanged throughout.)

\family-GCNet and \family-PSMNet have fewer parameters than their counterparts. Specifically, our networks have 2.3M and 1.9M parameters versus 2.6M and 5.2M of the baselines. Parallelized matching feature generation is efficient and takes 289 msec for a $256 \times 512$ input image. Therefore, network training is as fast as the baselines.

\begin{table*}[t]
\small
\centering
\centering
\scalebox{0.83}{
	\begin{tabular}{c|cccccccccc|cc}
	\multicolumn{2}{c}{}& \multicolumn{9}{c}{$\mathcal{S}$ source domain: sf-all} & \\
	\hline
	\makecell{$\mathcal{T}$ target \\ domain } & \makecell{MADNet$^1$ \\ \cite{Tonioni_2019_CVPR}} & \makecell{DispNet$^2$ \\ \cite{mayer2016large}} & \makecell{CRL$^2$ \\ \cite{pang2017cascade}} & \makecell{iResNet$^2$ \\ \cite{liang2018learning_1st_Rob}} & \makecell{SegStereo$^2$ \\ \cite{yang2018segstereo}} & \makecell{EdgeStereo$^2$ \\ \cite{song2019edgestereo}} & \makecell{GWC-Net$^3$ \\ \cite{guo2019group}} & \makecell{GANet$^3$ \\ \cite{zhang2019ga}} & \makecell{HD3$^3$ \\ \cite{Yin_2019_CVPR}} & \makecell{DSMNet$^3$ \\ \cite{zhang2019domain}} & \makecell{\family-\\GCNet} & \makecell{\family-\\PSMNet} \\ 
	\hline
	KT12 & 39.17 & 12.54 & 9.07 & 7.90 & 12.80 & 12.27 & 20.20 & 10.10 & 23.60 & 6.20 & \bfseries 5.51 & 13.97 \\
	KT15 & 43.98 & 12.88 & 8.88 & 7.42 & 11.23 & 12.47 & 22.70 & 11.70 & 26.50 & 6.50 & \bfseries 6.21 &  7.76 \\
	\hline
	
\end{tabular}
}
\vspace{-4pt}
\caption{\textbf{Comparison between \family-Nets and state-of-the-art 2D and 3D architectures.} All models are trained on \textit{sf-all} and tested on KITTI 2012 (top) and KITTI 2015 (bottom) training sets. Results obtained: $^1$using authors' weights, $^2$from \cite{song2019edgestereo} or $^3$ from \cite{zhang2019domain}. 
}
\label{tab:2dnets}
\end{table*}

\begin{table*}[t]
\small
\centering
\scalebox{0.85}{
\begin{tabular}{lc cccc| cccc}
&	&	\multicolumn{8}{c}{Target domain $\mathcal{T}$} \\
\cline{3-10} 
& & \multicolumn{4}{c|}{KT15 (bad3-all)\%} & \multicolumn{4}{c}{MB (bad2-noc)\%} \\
\hline
\multirow{4}{*}{\rotatebox{90}{Src domain $\mathcal{S}$}}
	& \multicolumn{1}{c|}{} & GCNet & \family-GCNet & PSMNet & \family-PMSNet & GCNet & \family-GCNet & PSMNet & \family-PMSNet \\ 
\cline{2-10} 

& \multicolumn{1}{r|}{sf-all} & 14.68 &\textbf{6.21}  &26.62 & \textbf{7.76} &30.42 & \textbf{18.52} & 26.92 &\textbf{19.81} \\
\cline{2-10}
& \multicolumn{1}{r|}{sf-all$\rightarrow$KT12} & 4.05 & \textbf{3.57} & \textbf{2.92} & 4.17 & 33.34 & \textbf{25.63} & 20.46 &\textbf{18.24} \\
& \multicolumn{1}{r|}{sf-all$\rightarrow$ETH3D} & 16.41 & \textbf{6.97} & 14.98 & \textbf{9.43} & 51.05 & \textbf{34.23} & 30.15 &\textbf{24.36} \\
& \multicolumn{1}{r|}{sf-all$\rightarrow$KT12+ETH3D} & \textbf{4.29} & 4.42 & \textbf{3.11} & 3.57 & 23.34 & \textbf{20.45} & 20.19 &\textbf{18.65} \\
 \hline
\end{tabular} 
} 
\vspace{-3pt}
\caption{\textbf{Results after specialization on real data for \family-Nets network, GCNet and PSM-Net.} The bad-$x$ errors are evaluated on $\mathcal {T}$ after training on $\mathcal {S}$ without further fine-tuning or adaptation to $\mathcal {T}$.
}
\label{tab:finetune}
\end{table*}

\subsection{Domain Transfer Evaluation}\label{sec:domain_transfer}

In this section, we carefully analyze the performance of our proposed \family-Nets in a domain transfer setting. The \textit{sf-all} 
entries in Table \ref{tab:scarce_labeled_data} (top row in each subtable) show the bad-$x$\footnote{$x$ is the default threshold specified by each dataset: i.e., 3 for KT12/15, 2 for MB, and 1 for ETH3D. The default setting of each benchmark regarding occlusion is also applied, i.e. occluded pixels are considered in KT15.} error for domain transfer of GCNet, PSMNet and their \family-Nets counterparts from Scene Flow to the other datasets. 

According to Table \ref{tab:kitti_benchmark}, PSMNet performs better than GCNet on the KITTI benchmark. However, in terms of generalization performance, in most cases GCNet is better than PSMNet (see also Table \ref{tab:scarce_labeled_data}). We argue that it is due to GCNet having fewer parameters in feature extraction and hence less vulnerability to overfitting to the RGB data.
The table highlights that in most cases \family-Nets are better when transferred without adaptation to different domains than GCNet and PSMNet. The bad-$x$ errors are evaluated on the target datasets without any fine-tuning or adaptation once trained on the source dataset. For fair comparison with the baselines, in order to test in the target domain, the model trained on the source domain is chosen just according to its performance on the validation set (still in the source domain). 

Inspecting the top row of each subtable reveals that 
\family-GCNet always outperforms GCNet except for \textit{sf-all} $\rightarrow$ \textit{ETH3D}, while \family-PSMNet is more accurate than PSMNet by a wide margin. 
This different performances between these two variants can be well explained by the fact that GCNet and PSMNet have distinct parameter configuration (see Section \ref{sec:approach} and \ref{sec:scare-labled-data}). These rows evaluate the case when ample training data, 35,000 stereo pairs with ground truth here, are available. In the next section, we investigate the effects of data scarcity. 

\subsection{Generalization on Scarce Labeled Data} \label{sec:scare-labled-data}

Most of the existing deep learning methods capture the patterns and regularities of the training domain and can make reliable predictions in new domains only after fine-tuning on a sufficient amount of labeled data from the target domain. 
We argue that by mixing in geometric constraints and prior knowledge, \family-Nets can capture general properties which are suitable to both the source $\mathcal{S}$ and target $\mathcal{T}$ domains, instead of being specific to the source domain $\mathcal{S}$. Therefore, we continue by analyzing the generalization of our approach with  scarce labeled data in the following settings of \textit{sf-3k}, \textit{sfD3k}, \textit{sfM3k},and  \textit{sfF3k}\footnote{\textit{sf-all}: all training data of SF; \textit{sf-3k}: 3k-image subset of \textit{SF}; \textit{sfD/F/M3k}: 3k-image subset of the Driving (\textit{sfD}), Monkaa (\textit{sfM}) or FlyingThings3D (\textit{sfF}) subsets.\label{note-sfX}}, shown in Table \ref{tab:scarce_labeled_data} after the first row 
of each subtable.

In general, the two 3D convolutional networks (and their variants) behave differently. 
In particular, PSMNet counts more parameters than GCNet (i.e., 3.3M versus 0.3M) in unary feature extraction, but less parameters (i.e., 1.9M versus 2.3M) in 3D CNN layers for cost matching regularization, thus their vulnerability to overfitting to the color space, and hence their capacity to generalize is much different, 
\eg PSMNet trained on \textit{sf-all} performs worse than GCNet on real images (Table \ref{tab:scarce_labeled_data}).

The accuracy of GCNet drops, in most cases, when trained on scarce labeled data in place of \textit{sf-all}.
%
Due to its high capacity in deep feature extraction, PSMNet cannot be trained effectively on approximately 12 times fewer training examples. This leads to large errors and unpredictable behavior. For example, PSMNet produces surprisingly accurate results for \textit{sfM3k} $\rightarrow$ \textit{KT12/KT15}, but performs poorly in most other combinations.

Comparing GCNet with \family-GCNet, we see that the latter is typically more accurate, often by a wide margin, with the exception of the ETH3D data, on which accuracy is unpredictable. 
\family-PSMNet outperforms PSMNet on the real test images with one exception on the ETH3D data. Moreover, \family-Nets exhibit more stable accuracy, which is desirable for potential deployment in the wild.

\iftrue
\begin{table}
\small
\centering
\scalebox{0.9}{
    \begin{tabularx}{\linewidth}{p{1.7cm} | XXl | XXX }
    \hline
	\multirow{2}{0.9pt}{{\bf Models}} & \multicolumn{3}{c|}{\bf All-D1 \%} & \multicolumn{3}{c}{ \bf Noc-D1 \%} \\
	\cline{2-7} 
	 &{ bg} & {fg} & {all} & { bg} & {fg} & {all} \\
	\hline \hline
	\family-GCNet & 2.58 & 6.83 & 3.29  & 2.19 & 5.59	& 2.75  \\
	GC-Net & 2.21 & 6.16 & 2.87 & 2.02 & 5.58 & 2.61 \\
	\hline
	\family-PSMNet & 2.15 & 5.01 & 2.63 & 1.99 & 4.52 & 2.41 \\
	PSM-Net & {\bf 1.86} & {\bf 4.62} & {\bf 2.32} & {\bf 1.71} &{\bf 4.31} & {\bf2.14} \\
	\hline 
	\end{tabularx}}
\vspace{-8pt}
\caption*{ \footnotesize (a) Test results on KITTI 2015 Benchmark}
\label{tab:kt15_bench}

\vspace{2pt}
\centering
\scalebox{0.9}{
    \begin{tabularx}{\linewidth}{l | Xl| Xl | XX }
    \hline
	\multirow{2}{2pt}{{\bf Models}} & \multicolumn{2}{c|}{ \bf $>$ 3 px \%} & \multicolumn{2}{c|}{ \bf $>$ 5 px \%} & \multicolumn{2}{c}{ \bf Avg px} \\
	\cline{2-7} 
	 &{noc} & {all} & {noc} & {all} & {noc} & {all} \\
	\hline \hline
	\family-GCNet & 2.33 & 3.41 & 1.41  & 2.09 & 0.8	&  1.0 \\
	GC-Net & 1.77 & 2.30 & 1.12 & 1.46 & 0.6 & 0.7\\
	\hline
	\family-PSMNet & 3.52 & 4.26 & 1.98  & 2.48 & 0.9 & 1.0 \\
	PSM-Net & {\bf 1.49} & {\bf 1.89} & {\bf 0.90} & {\bf 1.15} & {\bf 0.5} & {\bf 0.6} \\
	\hline
   \end{tabularx}
}
\vspace{-8pt}
\caption*{ \footnotesize (b) Test results on KITTI 2012 Benchmark}
\label{tab:kt12_bench}
\vspace{-8pt}
\caption{\textbf{Results on KITTI 2015 and KITTI 2012 benchmarks.} Comparison among GCNet, PSM-Net and \family{} variants.}
\label{tab:kitti_benchmark}
\end{table}
\fi

\begin{figure*}[t]
    \centering
    \renewcommand{\tabcolsep}{0.2pt}
    \scriptsize
    \begin{tabular}{ccccc}
        \includegraphics[width=0.2\textwidth]{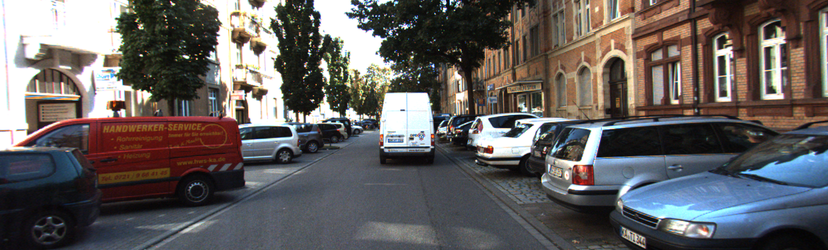} &  \includegraphics[width=0.2\textwidth]{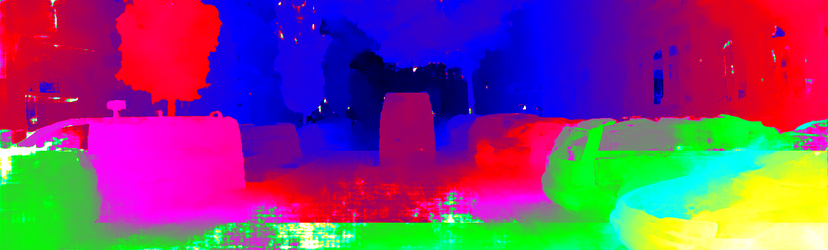} & \includegraphics[width=0.2\textwidth]{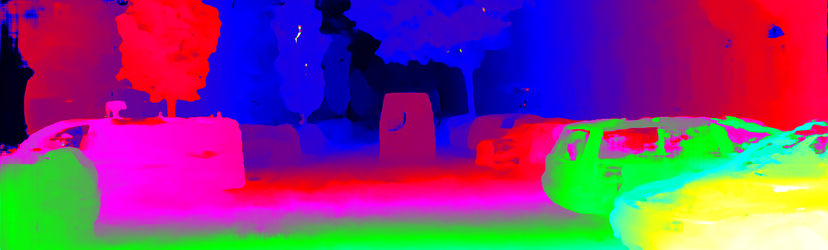} &
        \includegraphics[width=0.2\textwidth]{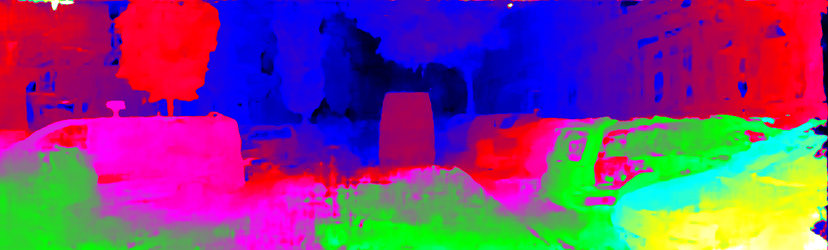} & \includegraphics[width=0.2\textwidth]{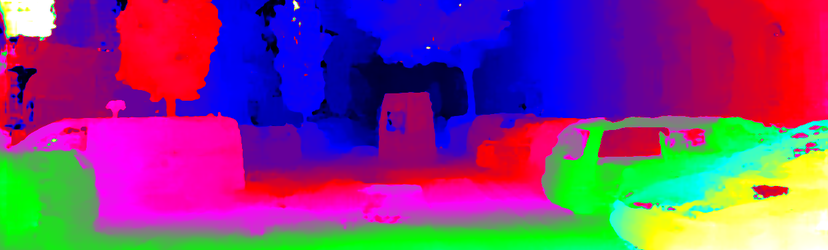} \\
        \includegraphics[width=0.2\textwidth]{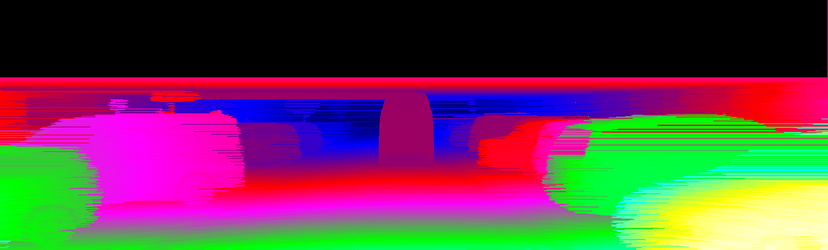} & 
        \begin{overpic}[width=0.2\textwidth]{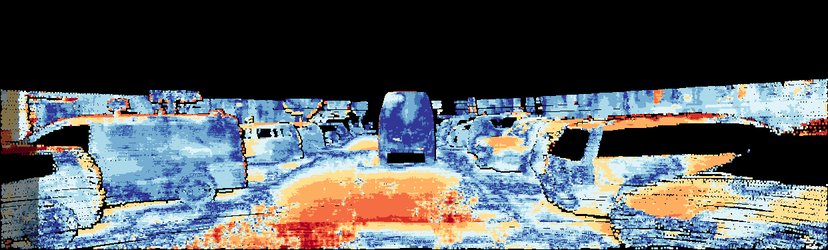}
        \put (4,22) {$\displaystyle\textcolor{white}{\textbf{26.85\%}}$}
        \end{overpic} & \begin{overpic}[width=0.2\textwidth]{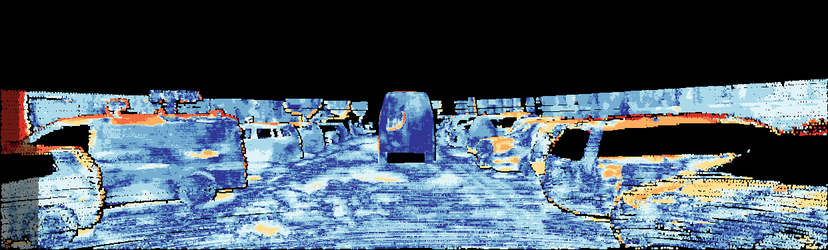}
        \put (4,22) {$\displaystyle\textcolor{white}{\textbf{8.59\%}}$}
        \end{overpic}
        &
        \begin{overpic}[width=0.2\textwidth]{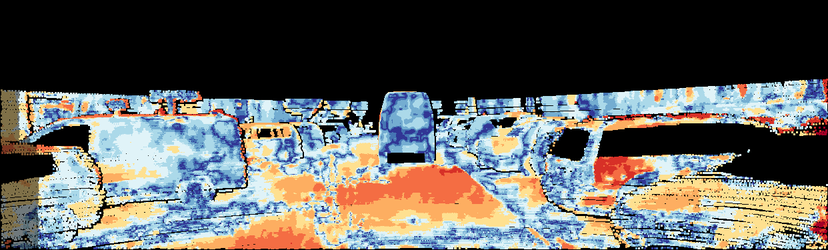}
        \put (4,22) {$\displaystyle\textcolor{white}{\textbf{36.46\%}}$}
        \end{overpic} & \begin{overpic}[width=0.2\textwidth]{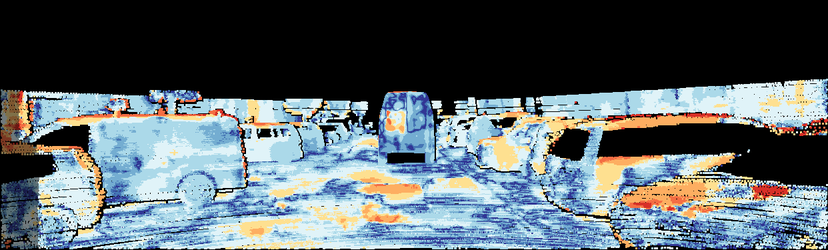}
        \put (4,22) {$\displaystyle\textcolor{white}{\textbf{14.71\%}}$}
        \end{overpic}
        \\
        \multicolumn{5}{c}{\includegraphics[width=\textwidth]{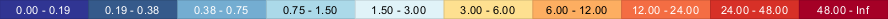}} \\
        \normalsize (a) & \normalsize (b) & \normalsize (c) & \normalsize (d) & \normalsize (e)\\
    \end{tabular}
    \vspace{-8pt}
    \caption{\textbf{Qualitative results on KITTI 2015 \cite{menze2015object} for networks trained on \textit{sf-all}.} Column (a) shows reference image and interpolated ground truth, (b) to (e) disparity (top) and error (bottom) maps obtained with GCNet \cite{kendall2017-gcnet}, \family-GCNet, PSMNet \cite{chang2018psmnet} and \family-PSMNet, respectively. Bad3-all rates are superimposed on the error maps.}
    \label{fig:qualitatives_kitti}
\end{figure*}

\begin{figure*}[t]
    \centering
    \renewcommand{\tabcolsep}{0.6pt}  
    \newcommand{\subfigwidth}{0.19}   
    \begin{tabular}{ccccc}
        \includegraphics[width=\subfigwidth\textwidth]{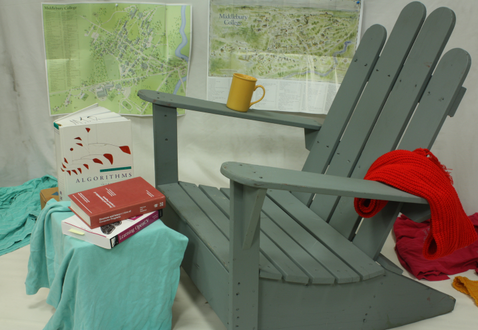} 
        & 
        \begin{overpic}[width=\subfigwidth\textwidth]{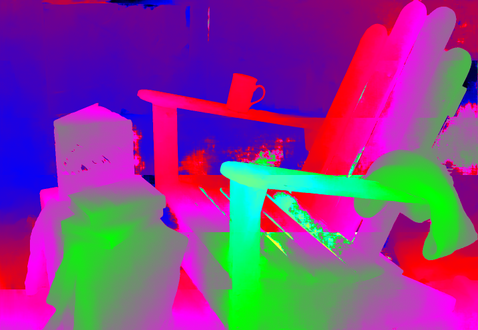}
        \put (4,55) {$\displaystyle\textcolor{white}{\textbf{30.44\%}}$}
        \end{overpic}
        &
        \begin{overpic}[width=\subfigwidth\textwidth]{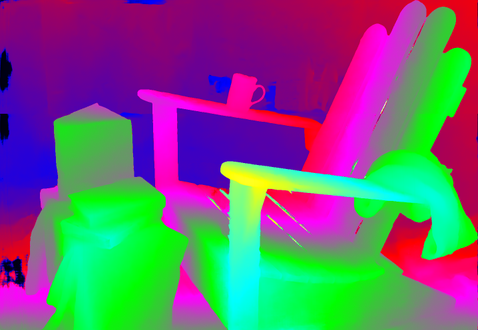}
        \put (4,55) {$\displaystyle\textcolor{white}{\textbf{11.10\%}}$}
        \end{overpic}
        &
        \begin{overpic}[width=\subfigwidth\textwidth]{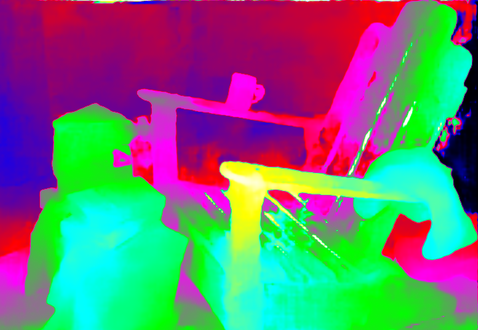}
        \put (4,55) {$\displaystyle\textcolor{white}{\textbf{31.40\%}}$}
        \end{overpic}
        &
        \begin{overpic}[width=\subfigwidth\textwidth]{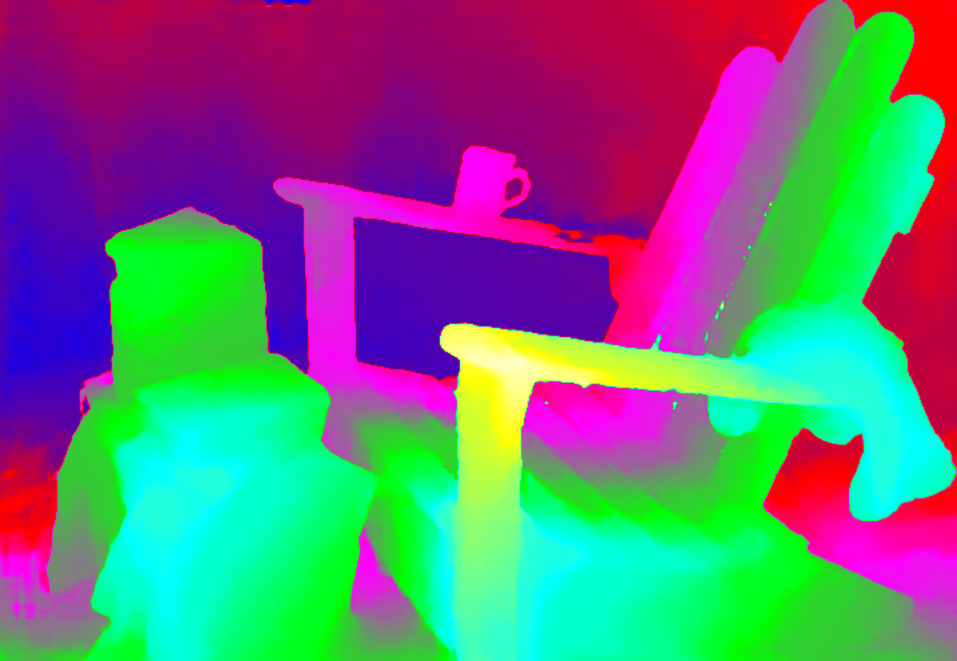}
        \put (4,55) {$\displaystyle\textcolor{white}{\textbf{10.38\%}}$}
        \end{overpic} 
        \\

        \includegraphics[width=\subfigwidth\textwidth]{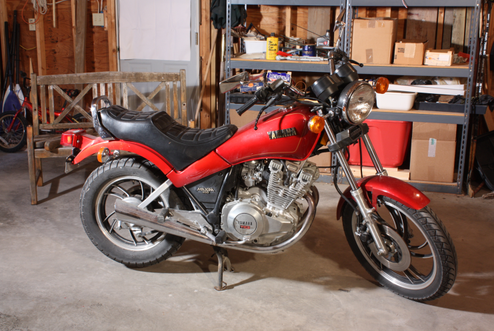} 
        & 
        \begin{overpic}[width=\subfigwidth\textwidth]{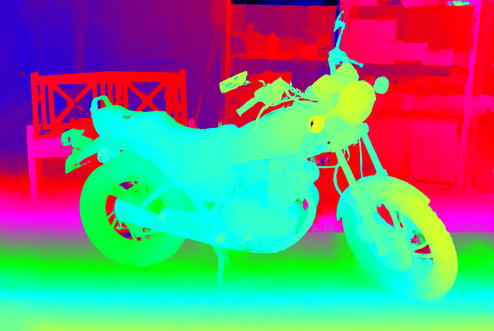}
        \put (4,55) {$\displaystyle\textcolor{white}{\textbf{19.81\%}}$}
        \end{overpic}
        &
        \begin{overpic}[width=\subfigwidth\textwidth]{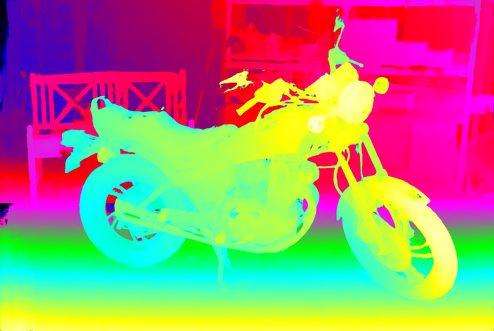}
        \put (4,55) {$\displaystyle\textcolor{white}{\textbf{12.39\%}}$}
        \end{overpic} 
        &
        \begin{overpic}[width=\subfigwidth\textwidth]{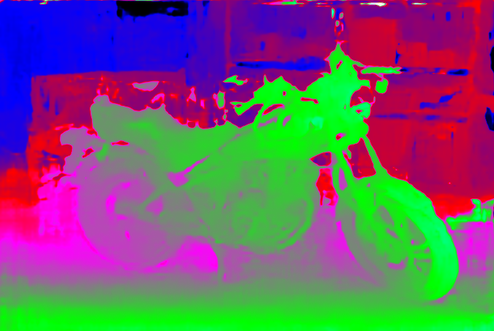}
        \put (4,55) {$\displaystyle\textcolor{white}{\textbf{46.02\%}}$}
        \end{overpic}
        &
        \begin{overpic}[width=\subfigwidth\textwidth]{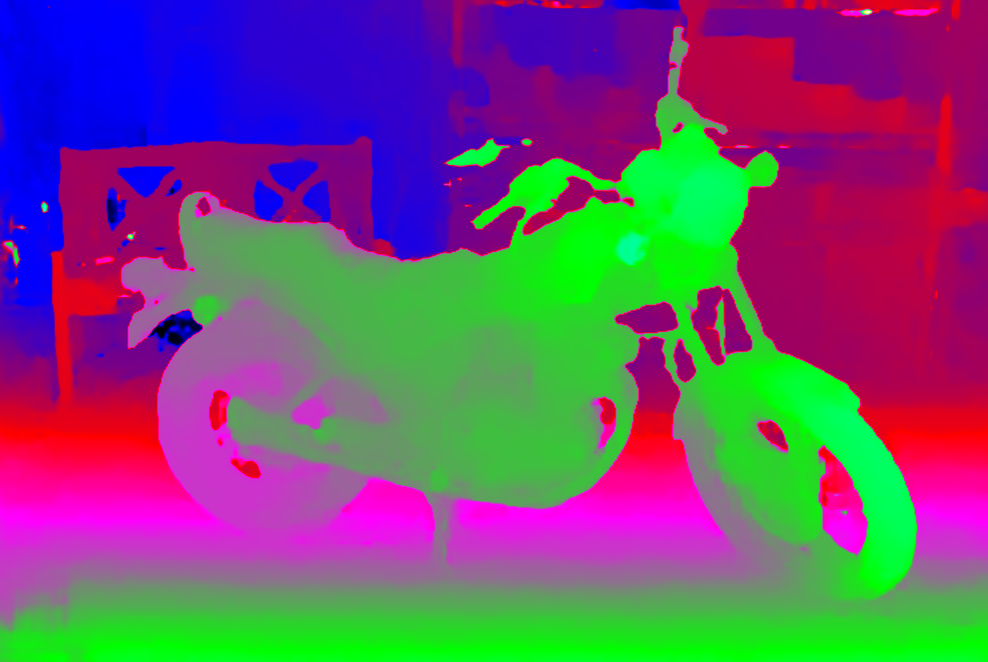}
        \put (4,55) {$\displaystyle\textcolor{white}{\textbf{14.57\%}}$}
        \end{overpic} 
        
        \\

    (a) & (b) & (c) & (d) & (e) \\
    \end{tabular}
    \vspace{-8pt}
    \caption{\textbf{Qualitative results on Middlebury 2014 \cite{scharstein2014high} for networks trained on \textit{sf-all}.} (a) Reference image, (b) to (e) disparity maps obtained with GCNet \cite{kendall2017-gcnet}, \family-GCNet, PSMNet \cite{chang2018psmnet} and \family-PSMNet, respectively. Bad2-noc rates are superimposed on the disparities.}
    \label{fig:qualitatives_middlebury}
\end{figure*}

\subsection{Comparison with State-of-the-art Networks}\label{sec:2d_arch}

In addition to a comparative study with respect to their direct 3D counterparts, we compare our \family-Nets with \emph{eleven 2D and 3D state-of-the-art networks} from the literature. To this aim, we follow the protocol of \cite{song2019edgestereo} and compute the accuracy on KT12 and KT15 after training on SceneFlow dataset (sf-all). 
Table \ref{tab:2dnets} summarizes these comparisons. 
\family-GCNet proves to be the best architecture at generalization in both domains, outperforming models that makes use of aggressive augmentation strategies (e.g., iResNet)  as well as concurrent proposals based on domain-normalization \cite{zhang2019domain}. This confirms that learning disparity estimation in Matching Space enables better generalization in totally unseen environments compared to carrying out the learning process in RGB space.
\family-PSMNet is often outperformed on KT12, while it is consistently more effective than eight out of ten baselines on KT15, as already observed in the previous experiments.

\subsection{Generalization across Real Domains}

In the next set of experiments, reported in Table \ref{tab:finetune}, we investigated the generalization of the networks pre-trained on synthetic data and then specialized on real data from a different domain. Specifically, we pre-trained all networks on \textit{sf-all} and then continued training for 300 more epochs on \textit{KT12}, \textit{ETH3D} or both datasets.


One of the key observations from Table \ref{tab:finetune} is that GCNet and PSMNet perform well on KT15, when data from KT12 are included in their training, but perform poorly otherwise (see column 1 and 3). \family-Nets have more stable performance even when the real domain used for specialization changes. GCNet and PSMNet perform poorly on MB since all the real domains on which they were trained have dissimilar appearance to it. On the other hand, the \family-Nets achieve higher accuracy on MB, considering that most end-to-end networks fail on this dataset.

\subsection{Evaluation on Source Domain Benchmark}

It is critical to see our \family-Nets not only show better generalization to unseen domains, but also obtain comparable performance to their counterparts, GCNet and PSMNet, on the test set of the source domains. Following the same protocol as GCNet and PSMNet, our models are firstly trained on synthetic data and fine-tuned on the KITTI 2015 and KITTI 2012 training sets, before being evaluated on the respective test sets. 
Table \ref{tab:kitti_benchmark} and the KITTI leaderboards show the test errors, with virtually no accuracy loss on KITTI 2015 and a moderate drop on KITTI 2012.
These results demonstrate that the cost of obtaining a network that generalizes well is small in terms of loss in specialization.

\subsection{Qualitative Results}
Figure \ref{fig:qualitatives_kitti} depicts an example from KITTI 2015 following the protocol of Section \ref{sec:domain_transfer}. \family-GCNet and \family-PSMNet produce almost smooth disparities on planar surfaces such as roads, whereas PSMNet and GCNet fail at handling the shift from the synthetic domain. For more examples in outdoor environments, we refer readers to the supplementary video sequences comparing GCNet with \family-GCNet and PSMNet with \family-PSMNet on KITTI raw sequence \textit{2011\_10\_03\_drive\_0034\_sync}.
Figure \ref{fig:qualitatives_middlebury} shows how \family-style models are much more accurate on Middlebury, better preserving in both cases the overall structure of the scene. For example, \family-PSMNet can recover small details such as the cup on top of the armrest (top row) or the front wheel of the motorcycle (bottom row) which are lost by PSMNet.

\section{Conclusions}\label{sec:conclusions}
We have introduced a novel family of 3D convolutional architectures for dense stereo matching, namely \family-Nets. 
By learning to reason on relationships in matching space without directly being affected by image appearance, our models show superior generalization to unseen domains. 
The matching space encodes invariant properties that have been proven to be effective in conventional stereo matching. Deep networks such as PSMNet, on the other hand, are purely data-driven and are unable to learn all relevant principles from their training data. Our approach strikes a balance between the empirical modeling and data-driven learning.
Conversely to known approaches for tackling the domain shift problem \cite{pang2018zoom-and-learn,tonioni2017unsupervised,Tonioni_2019_CVPR,zhong2018open}, our method can be transferred without the need for retraining or adaptation to new domains, thus presenting an appealing alternative for deployment in real applications. 

Our experiments on two state-of-the-art 3D convolutional architectures, GCNet and PSMNet, 
and their respective \family{} counterparts, trained on different amounts of synthetic data, confirm that \family-Nets generalize better to different realistic image context, both indoors and outdoors. 
Additional experiments comparing the generalization performance of \family-Nets with that of ten additional state-of-the-art CNNs for stereo matching show that \family-GCNet is superior in accuracy on unseen data  (Section \ref{sec:2d_arch}).










\textbf{Acknowledgements.} \,\, This research has been partially supported by National Science Foundation under Awards IIS-1527294 and IIS-1637761. We gratefully acknowledge the support of NVIDIA Corporation with the donation of the Titan Xp GPU used for this research.

{\small
\bibliographystyle{ieee}
\bibliography{ref}
}

\newpage
\begin{center}
    \section*{\fontsize{13}{15}\selectfont Supplement}
\end{center}

 
In this supplementary document, we present additional qualitative results on domain generalization (i.e., \textit{sf-all $\rightarrow$ others}), which are omitted in the main paper due to space limits. 

\begin{figure*}[hbt!]
    \centering
    \renewcommand{\tabcolsep}{0.2pt}
    \scriptsize
    \begin{tabular}{ccccc}
    
        \includegraphics[width=0.2\textwidth]{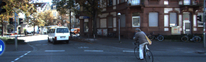} &  \includegraphics[width=0.2\textwidth]{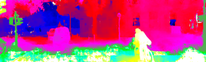} & \includegraphics[width=0.2\textwidth]{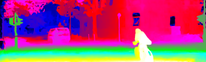} &
        \includegraphics[width=0.2\textwidth]{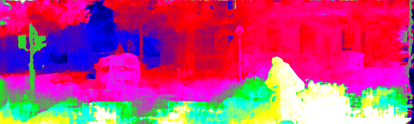} & \includegraphics[width=0.2\textwidth]{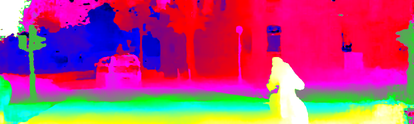} \\
        \includegraphics[width=0.2\textwidth]{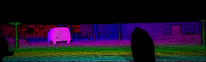} & 
        \begin{overpic}[width=0.2\textwidth]{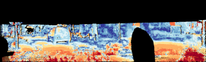}
        \put (4,22) {$\displaystyle\textcolor{white}{\textbf{38.57\%}}$}
        \end{overpic} & \begin{overpic}[width=0.2\textwidth]{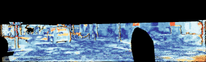}
        \put (4,22) {$\displaystyle\textcolor{white}{\textbf{6.16\%}}$}
        \end{overpic}
        &
        \begin{overpic}[width=0.2\textwidth]{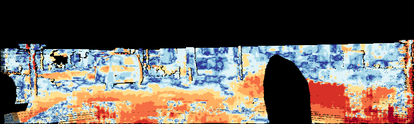}
        \put (4,22) {$\displaystyle\textcolor{white}{\textbf{40.51\%}}$}
        \end{overpic} & \begin{overpic}[width=0.2\textwidth]{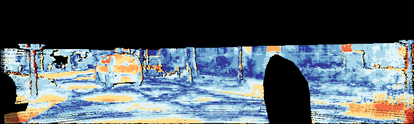}
        \put (4,22) {$\displaystyle\textcolor{white}{\textbf{15.93\%}}$}
        \end{overpic}
        \\
        
        \includegraphics[width=0.2\textwidth]{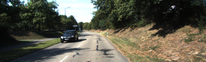} &  \includegraphics[width=0.2\textwidth]{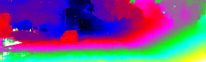} & \includegraphics[width=0.2\textwidth]{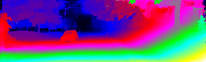} &
        \includegraphics[width=0.2\textwidth]{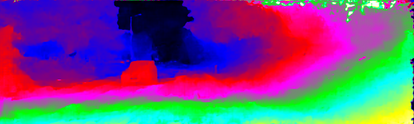} & \includegraphics[width=0.2\textwidth]{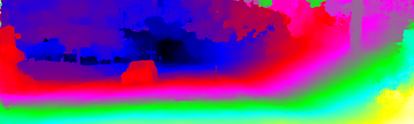} \\
        \includegraphics[width=0.2\textwidth]{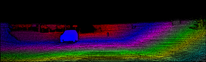} & 
        \begin{overpic}[width=0.2\textwidth]{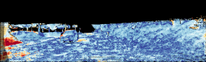}
        \put (4,22) {$\displaystyle\textcolor{white}{\textbf{3.04\%}}$}
        \end{overpic} & \begin{overpic}[width=0.2\textwidth]{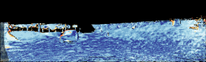}
        \put (4,22) {$\displaystyle\textcolor{white}{\textbf{1.21\%}}$}
        \end{overpic}
        &
        \begin{overpic}[width=0.2\textwidth]{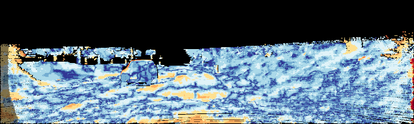}
        \put (4,22) {$\displaystyle\textcolor{white}{\textbf{8.33\%}}$}
        \end{overpic} & \begin{overpic}[width=0.2\textwidth]{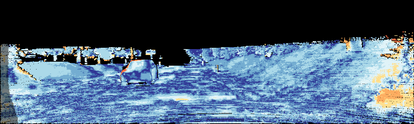}
        \put (4,22) {$\displaystyle\textcolor{white}{\textbf{3.05\%}}$}
        \end{overpic}
        \\
        
        \includegraphics[width=0.2\textwidth]{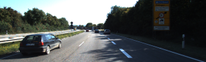} &  \includegraphics[width=0.2\textwidth]{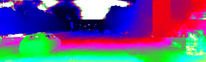} & \includegraphics[width=0.2\textwidth]{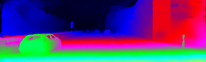} &
        \includegraphics[width=0.2\textwidth]{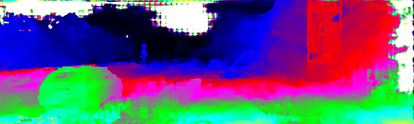} & \includegraphics[width=0.2\textwidth]{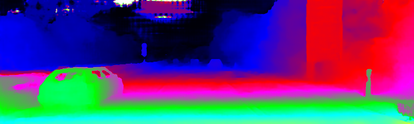} \\
        \includegraphics[width=0.2\textwidth]{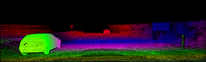} & 
        \begin{overpic}[width=0.2\textwidth]{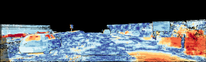}
        \put (4,22) {$\displaystyle\textcolor{white}{\textbf{23.98\%}}$}
        \end{overpic} & \begin{overpic}[width=0.2\textwidth]{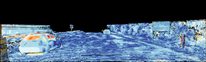}
        \put (4,22) {$\displaystyle\textcolor{white}{\textbf{4.21\%}}$}
        \end{overpic}
        &
        \begin{overpic}[width=0.2\textwidth]{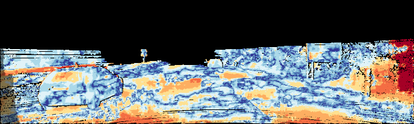}
        \put (4,22) {$\displaystyle\textcolor{white}{\textbf{25.46\%}}$}
        \end{overpic} & \begin{overpic}[width=0.2\textwidth]{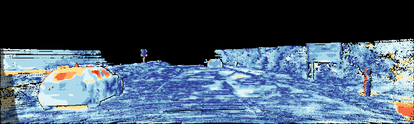}
        \put (4,22) {$\displaystyle\textcolor{white}{\textbf{5.33\%}}$}
        \end{overpic}
        \\
        
        \includegraphics[width=0.2\textwidth]{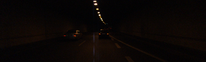} &  \includegraphics[width=0.2\textwidth]{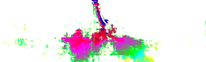} & \includegraphics[width=0.2\textwidth]{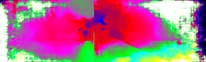} &
        \includegraphics[width=0.2\textwidth]{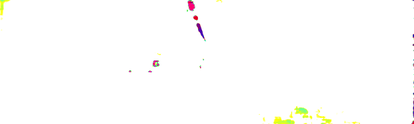} & \includegraphics[width=0.2\textwidth]{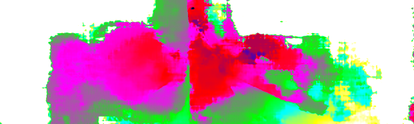} \\
        \includegraphics[width=0.2\textwidth]{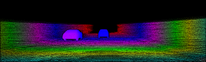} & 
        \begin{overpic}[width=0.2\textwidth]{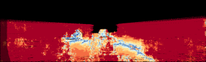}
        \put (4,22) {$\displaystyle\textcolor{white}{\textbf{94.16\%}}$}
        \end{overpic} & \begin{overpic}[width=0.2\textwidth]{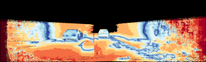}
        \put (4,22) {$\displaystyle\textcolor{white}{\textbf{61.03\%}}$}
        \end{overpic}
        &
        \begin{overpic}[width=0.2\textwidth]{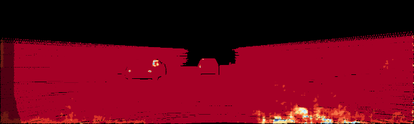}
        \put (4,22) {$\displaystyle\textcolor{white}{\textbf{99.69\%}}$}
        \end{overpic} & \begin{overpic}[width=0.2\textwidth]{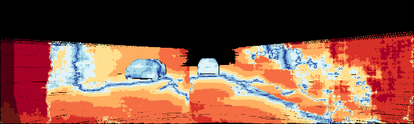}
        \put (4,22) {$\displaystyle\textcolor{white}{\textbf{73.08\%}}$}
        \end{overpic}
        \\
    
        \includegraphics[width=0.2\textwidth]{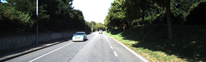} &  \includegraphics[width=0.2\textwidth]{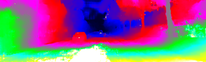} & \includegraphics[width=0.2\textwidth]{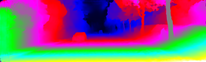} &
        \includegraphics[width=0.2\textwidth]{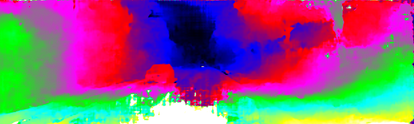} & \includegraphics[width=0.2\textwidth]{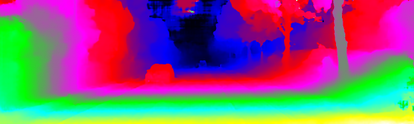} \\
        \includegraphics[width=0.2\textwidth]{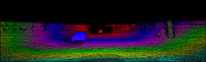} & 
        \begin{overpic}[width=0.2\textwidth]{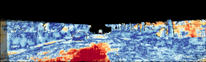}
        \put (4,22) {$\displaystyle\textcolor{white}{\textbf{22.42\%}}$}
        \end{overpic} & \begin{overpic}[width=0.2\textwidth]{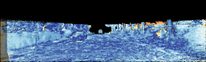}
        \put (4,22) {$\displaystyle\textcolor{white}{\textbf{1.78\%}}$}
        \end{overpic}
        &
        \begin{overpic}[width=0.2\textwidth]{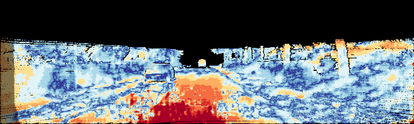}
        \put (4,22) {$\displaystyle\textcolor{white}{\textbf{24.66\%}}$}
        \end{overpic} & \begin{overpic}[width=0.2\textwidth]{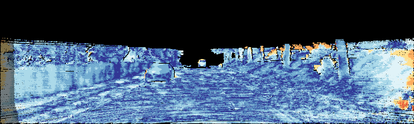}
        \put (4,22) {$\displaystyle\textcolor{white}{\textbf{2.42\%}}$}
        \end{overpic}
        \\
    
        \includegraphics[width=0.2\textwidth]{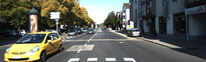} &  \includegraphics[width=0.2\textwidth]{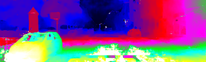} & \includegraphics[width=0.2\textwidth]{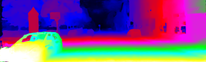} &
        \includegraphics[width=0.2\textwidth]{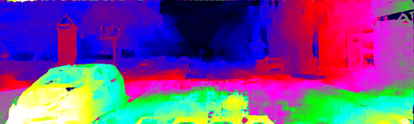} & \includegraphics[width=0.2\textwidth]{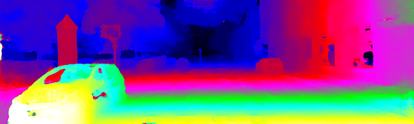} \\
        \includegraphics[width=0.2\textwidth]{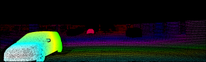} & 
        \begin{overpic}[width=0.2\textwidth]{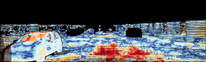}
        \put (4,22) {$\displaystyle\textcolor{white}{\textbf{24.07\%}}$}
        \end{overpic} & \begin{overpic}[width=0.2\textwidth]{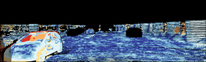}
        \put (4,22) {$\displaystyle\textcolor{white}{\textbf{13.12\%}}$}
        \end{overpic}
        &
        \begin{overpic}[width=0.2\textwidth]{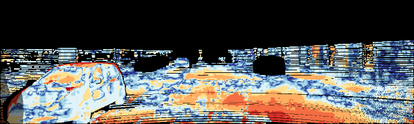}
        \put (4,22) {$\displaystyle\textcolor{white}{\textbf{29.43\%}}$}
        \end{overpic} & \begin{overpic}[width=0.2\textwidth]{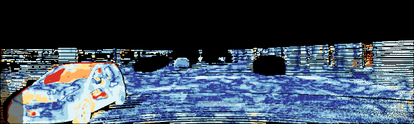}
        \put (4,22) {$\displaystyle\textcolor{white}{\textbf{13.83\%}}$}
        \end{overpic}
        \\
        
        \includegraphics[width=0.2\textwidth]{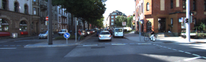} &  \includegraphics[width=0.2\textwidth]{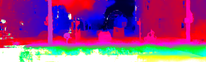} & \includegraphics[width=0.2\textwidth]{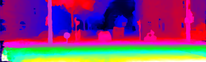} &
        \includegraphics[width=0.2\textwidth]{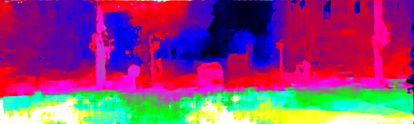} & \includegraphics[width=0.2\textwidth]{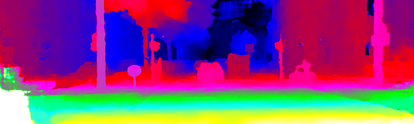} \\
        \includegraphics[width=0.2\textwidth]{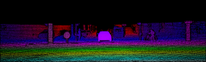} & 
        \begin{overpic}[width=0.2\textwidth]{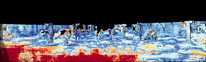}
        \put (4,22) {$\displaystyle\textcolor{white}{\textbf{32.67\%}}$}
        \end{overpic} & \begin{overpic}[width=0.2\textwidth]{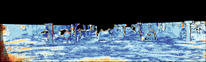}
        \put (4,22) {$\displaystyle\textcolor{white}{\textbf{6.80\%}}$}
        \end{overpic}
        &
        \begin{overpic}[width=0.2\textwidth]{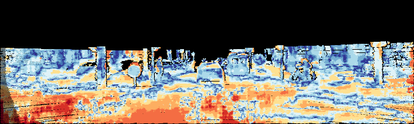}
        \put (4,22) {$\displaystyle\textcolor{white}{\textbf{39.06\%}}$}
        \end{overpic} & \begin{overpic}[width=0.2\textwidth]{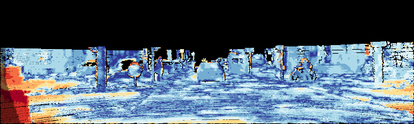}
        \put (4,22) {$\displaystyle\textcolor{white}{\textbf{9.49\%}}$}
        \end{overpic}
        \\
        
        \includegraphics[width=0.2\textwidth]{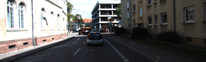} &  \includegraphics[width=0.2\textwidth]{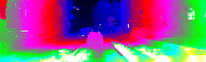} & \includegraphics[width=0.2\textwidth]{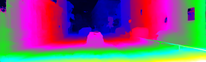} &
        \includegraphics[width=0.2\textwidth]{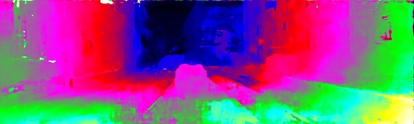} & \includegraphics[width=0.2\textwidth]{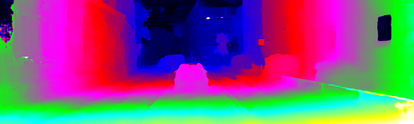} \\
        \includegraphics[width=0.2\textwidth]{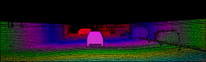} & 
        \begin{overpic}[width=0.2\textwidth]{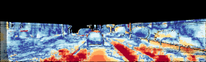}
        \put (4,22) {$\displaystyle\textcolor{white}{\textbf{28.87\%}}$}
        \end{overpic} & \begin{overpic}[width=0.2\textwidth]{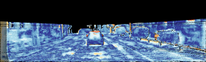}
        \put (4,22) {$\displaystyle\textcolor{white}{\textbf{2.46\%}}$}
        \end{overpic}
        &
        \begin{overpic}[width=0.2\textwidth]{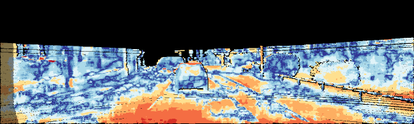}
        \put (4,22) {$\displaystyle\textcolor{white}{\textbf{28.28\%}}$}
        \end{overpic} & \begin{overpic}[width=0.2\textwidth]{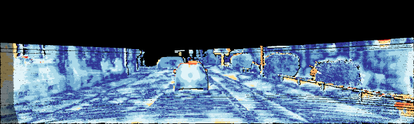}
        \put (4,22) {$\displaystyle\textcolor{white}{\textbf{3.23\%}}$}
        \end{overpic}
        \\
        
        \includegraphics[width=0.2\textwidth]{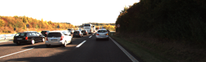} &  \includegraphics[width=0.2\textwidth]{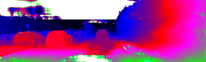} & \includegraphics[width=0.2\textwidth]{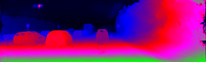} &
        \includegraphics[width=0.2\textwidth]{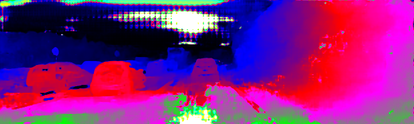} & \includegraphics[width=0.2\textwidth]{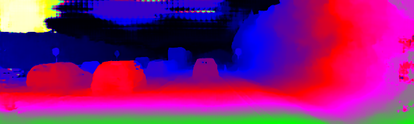} \\
        \includegraphics[width=0.2\textwidth]{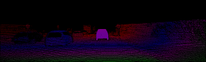} & 
        \begin{overpic}[width=0.2\textwidth]{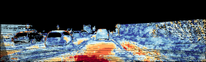}
        \put (4,22) {$\displaystyle\textcolor{white}{\textbf{26.99\%}}$}
        \end{overpic} & \begin{overpic}[width=0.2\textwidth]{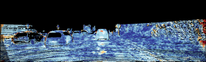}
        \put (4,22) {$\displaystyle\textcolor{white}{\textbf{3.64\%}}$}
        \end{overpic}
        &
        \begin{overpic}[width=0.2\textwidth]{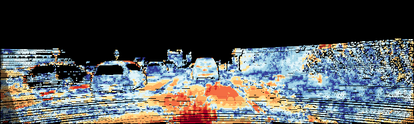}
        \put (4,22) {$\displaystyle\textcolor{white}{\textbf{27.80\%}}$}
        \end{overpic} & \begin{overpic}[width=0.2\textwidth]{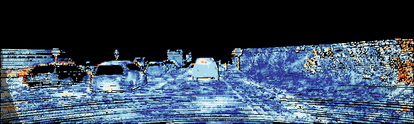}
        \put (4,22) {$\displaystyle\textcolor{white}{\textbf{3.06\%}}$}
        \end{overpic}
        \\
        
        \multicolumn{5}{c}{\includegraphics[width=\textwidth]{imgs/qualitatives/kitti_scale.png}} \\
        \normalsize (a) & \normalsize (b) & \normalsize (c) & \normalsize (d) & \normalsize (e)\\
    \end{tabular}
    \vspace{-8pt}
    \caption{Qualitative results on KITTI 2015 \cite{menze2015object} for networks trained on \textit{sf-all}. Every two rows correspond to one example in KITTI 2015 training set. (a) shows reference image and ground truth, (b) to (e) disparity (top) and error (bottom) maps obtained with GCNet \cite{kendall2017-gcnet}, \family-GCNet, PSMNet \cite{chang2018psmnet} and \family-PSMNet, respectively. Bad3-all rates are superimposed on the error maps.}
    \label{fig:supp-qualitatives_kt15}
\end{figure*}

\vspace{8pt}\noindent \textbf{Qualitative results from sf-all $\rightarrow$ KT15} \,\, In Fig. \ref{fig:supp-qualitatives_kt15} we provide more qualitative results on KITTI 2015 (KT15) \cite{menze2015object} for the networks trained on \textit{sf-all} \cite{mayer2016large}. Every two rows correspond to an example from the KT15 training set. Specifically, Fig. \ref{fig:supp-qualitatives_kt15}(a) shows the reference image and ground truth, Fig. \ref{fig:supp-qualitatives_kt15}(b) to (e) show the disparity (top) and error (bottom) maps obtained by baseline GCNet \cite{kendall2017-gcnet}, our \family-GCNet, baseline PSMNet \cite{chang2018psmnet} and our \family-PSMNet, respectively. Please note the bad3-all rates are superimposed on the error maps.

\begin{figure*}[hbt!]
    \centering
    \renewcommand{\tabcolsep}{0.2pt} 
    \begin{tabular}{ccccc}
\iftrue
        \includegraphics[width=0.2\textwidth]{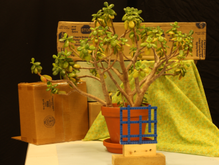} 
        & 
        \begin{overpic}[width=0.2\textwidth]{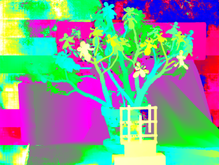}
        \put (4,55) {$\displaystyle\textcolor{black}{\textbf{48.47\%}}$}
        \end{overpic}
        &
        \begin{overpic}[width=0.2\textwidth]{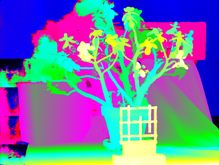}
        \put (4,55) {$\displaystyle\textcolor{black}{\textbf{20.10\%}}$}
        \end{overpic}
        &
        \begin{overpic}[width=0.2\textwidth]{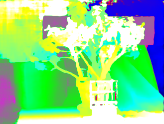} 
        \put (4,55) {$\displaystyle\textcolor{black}{\textbf{58.63\%}}$}
        \end{overpic}
        &
        \begin{overpic}[width=0.2\textwidth]{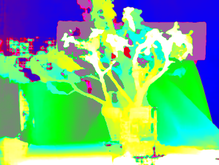}
        \put (4,55) {$\displaystyle\textcolor{black}{\textbf{52.72\%}}$}
        \end{overpic} 
        \\
\fi
\iftrue
        \includegraphics[width=0.2\textwidth]{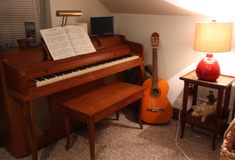} 
        & 
        \begin{overpic}[width=0.2\textwidth]{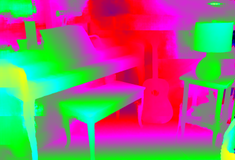}
        \put (4,55) {$\displaystyle\textcolor{black}{\textbf{31.78\%}}$}
        \end{overpic}
        &
        \begin{overpic}[width=0.2\textwidth]{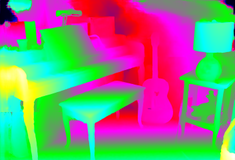}
        \put (4,55) {$\displaystyle\textcolor{black}{\textbf{19.91\%}}$}
        \end{overpic}
        &
        \begin{overpic}[width=0.2\textwidth]{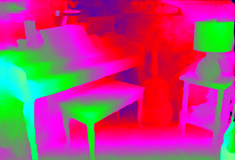}
        \put (4,55) {$\displaystyle\textcolor{black}{\textbf{28.34\%}}$}
        \end{overpic}
        &
        \begin{overpic}[width=0.2\textwidth]{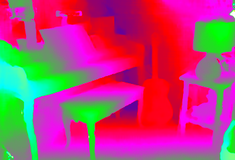}
        \put (4,55) {$\displaystyle\textcolor{black}{\textbf{23.34\%}}$}
        \end{overpic} 
         \\
\fi

\iftrue
        \includegraphics[width=0.2\textwidth]{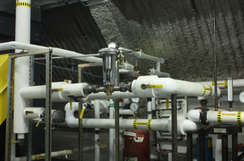} 
        & 
        \begin{overpic}[width=0.2\textwidth]{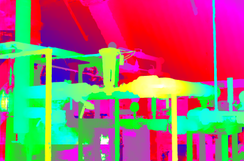}
        \put (10,50) {$\displaystyle\textcolor{black}{\textbf{20.39\%}}$}
        \end{overpic}
        &
        \begin{overpic}[width=0.2\textwidth]{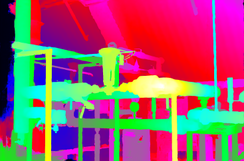}
        \put (10,50) {$\displaystyle\textcolor{black}{\textbf{10.86\%}}$}
        \end{overpic} 
        &
        \begin{overpic}[width=0.2\textwidth]{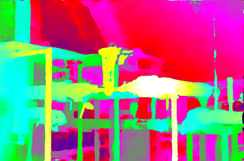}
        \put (10,50) {$\displaystyle\textcolor{black}{\textbf{36.18\%}}$}
        \end{overpic}
        &
        \begin{overpic}[width=0.2\textwidth]{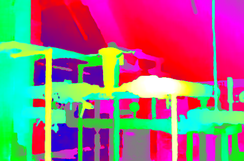}
        \put (10,50) {$\displaystyle\textcolor{black}{\textbf{26.42\%}}$}
        \end{overpic} 
        \\
\fi

\iftrue
        \includegraphics[width=0.2\textwidth]{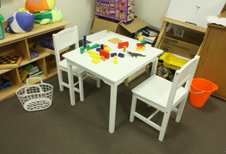} 
        & 
        \begin{overpic}[width=0.2\textwidth]{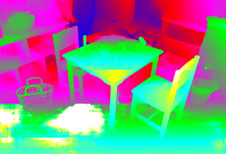}
        \put (4,55) {$\displaystyle\textcolor{black}{\textbf{35.86\%}}$}
        \end{overpic}
        &
        \begin{overpic}[width=0.2\textwidth]{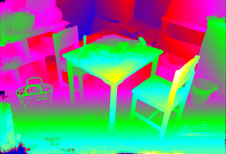}
        \put (4,55) {$\displaystyle\textcolor{black}{\textbf{19.0\%}}$}
        \end{overpic} 
        & 
       \begin{overpic}[width=0.2\textwidth]{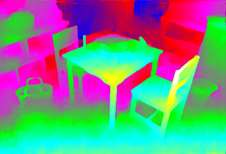}
        \put (4,55) {$\displaystyle\textcolor{black}{\textbf{32.56\%}}$}
        \end{overpic}
        &
        \begin{overpic}[width=0.2\textwidth]{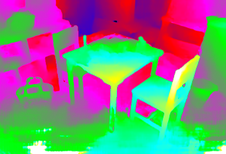}
        \put (4,55) {$\displaystyle\textcolor{black}{\textbf{43.69\%}}$}
        \end{overpic}
        \\
\fi  

\iftrue
        \includegraphics[width=0.2\textwidth]{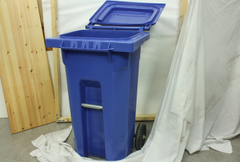} 
        & 
        \begin{overpic}[width=0.2\textwidth]{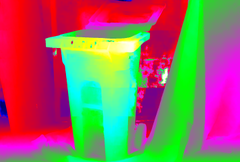}
        \put (4,55) {$\displaystyle\textcolor{black}{\textbf{25.08\%}}$}
        \end{overpic}
        &
        \begin{overpic}[width=0.2\textwidth]{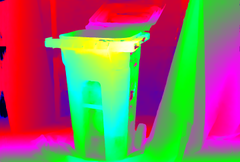}
        \put (4,55) {$\displaystyle\textcolor{black}{\textbf{13.51\%}}$}
        \end{overpic} 
        & 
       \begin{overpic}[width=0.2\textwidth]{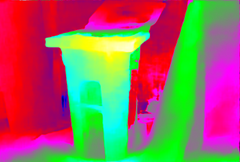}
        \put (4,55) {$\displaystyle\textcolor{black}{\textbf{25.52\%}}$}
        \end{overpic}
        &
        \begin{overpic}[width=0.2\textwidth]{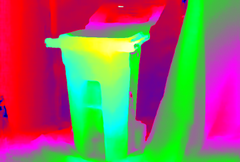}
        \put (4,55) {$\displaystyle\textcolor{black}{\textbf{13.84\%}}$}
        \end{overpic}
        \\
\fi  

\iftrue
        \includegraphics[width=0.2\textwidth]{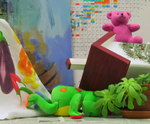} 
        & 
        \begin{overpic}[width=0.2\textwidth]{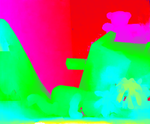}
        \put (10,70) {$\displaystyle\textcolor{black}{\textbf{20.62\%}}$}
        \end{overpic}
        &
        \begin{overpic}[width=0.2\textwidth]{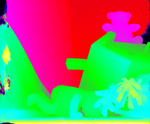}
        \put (10,70) {$\displaystyle\textcolor{black}{\textbf{5.07\%}}$}
        \end{overpic} 
        & 
       \begin{overpic}[width=0.2\textwidth]{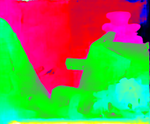}
        \put (10,70) {$\displaystyle\textcolor{black}{\textbf{26.49\%}}$}
        \end{overpic}
        &
        \begin{overpic}[width=0.2\textwidth]{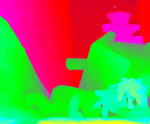}
        \put (10,70) {$\displaystyle\textcolor{black}{\textbf{15.39\%}}$}
        \end{overpic}
        \\
\fi  
    (a) & (b) & (c) & (d) & (e) \\
    \end{tabular}
    \vspace{-8pt}
    \caption{Qualitative results on Middlebury 2014 \cite{scharstein2014high} for networks trained on \textit{sf-all}. Column (a) shows reference image, (b) to (e) disparity maps obtained with GCNet \cite{kendall2017-gcnet}, \family-GCNet, PSMNet \cite{chang2018psmnet} and \family-PSMNet, respectively. Bad2-noc rates are superimposed on the disparity maps.}
    \label{fig:supp_qualitatives_middlebury}
\end{figure*}

\vspace{8pt}\noindent \textbf{Qualitative results from sf-all $\rightarrow$ MB} \,\, Fig. \ref{fig:supp_qualitatives_middlebury} shows additional qualitative results on Middlebury 2014 (MB) \cite{scharstein2014high} for the networks trained on \textit{sf-all}. Each row corresponds to an example from the MB training set. Fig. \ref{fig:supp_qualitatives_middlebury}(a) shows the reference image, Fig. \ref{fig:supp_qualitatives_middlebury}(b) to (e) are the disparity maps obtained by baseline GCNet \cite{kendall2017-gcnet}, our \family-GCNet, baseline PSMNet \cite{chang2018psmnet} and our \family-PSMNet, respectively. Bad2-noc rates are superimposed on the disparity maps.

\vspace{8pt}\noindent \textbf{Qualitative results from sf-all $\rightarrow$ KT Raw} \,\, We show more qualitative results on KITTI (KT) raw sequence \textit{2011\_10\_03\_drive\_0034\_sync}, for \family{}-GCNet (Fig. \ref{fig:supp-kt-raw-msgcnet}) and \family{}-PSMNet (Fig. \ref{fig:supp-kt-raw-mspsmnet}) trained on \textit{sf-all}. Specifically, Fig. \ref{fig:supp-kt-raw-msgcnet}(a) is the input left frame, and Fig. \ref{fig:supp-kt-raw-msgcnet}(b) and (c) are the disparity maps estimated by baseline GCNet \cite{kendall2017-gcnet} and our \family-GCNet, respectively. Fig \ref{fig:supp-kt-raw-mspsmnet} provides the results for baseline PSMNet \cite{chang2018psmnet} and our \family{}-PSMNet. Please note there is no ground truth for KT raw sequences, so the error rates cannot be calculated. Still, comparing the disparity maps in (b) and (c), ours tend to predict more reliable and smooth disparities rather than the noisy and bumpy ones by the baselines. For more examples, please see 
our \family{}-GCNet video (\url{https://youtu.be/Qr6WGsPX5P8}) and \family{}-PSMNet video (\url{https://youtu.be/t9WPc3pxzc4}). Each frame of the videos shows the input left image (top), the disparity map estimated by the baseline (middle), and the disparity map by our \family{} counterpart (bottom).

\iftrue
\begin{figure*}[hbt!]
    \centering
    \renewcommand{\tabcolsep}{1pt}
    \scriptsize
    \begin{tabular}{ccc}
        \includegraphics[width=0.32\textwidth]{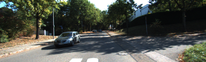} &
        \includegraphics[width=0.32\textwidth]{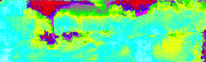} &
        \includegraphics[width=0.32\textwidth]{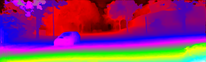} \\
        
        \includegraphics[width=0.32\textwidth]{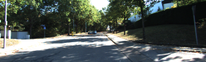} &
        \includegraphics[width=0.32\textwidth]{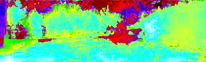} &
        \includegraphics[width=0.32\textwidth]{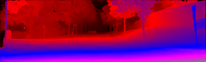} 

        \\
        
        \includegraphics[width=0.32\textwidth]{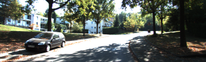} &
        \includegraphics[width=0.32\textwidth]{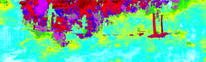} &
        \includegraphics[width=0.32\textwidth]{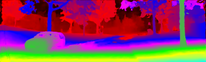} 
        
        \\
        
        \includegraphics[width=0.32\textwidth]{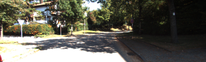} &
        \includegraphics[width=0.32\textwidth]{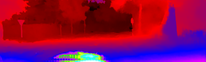} &
        \includegraphics[width=0.32\textwidth]{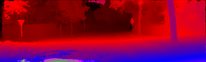} 
  \\
        
        \includegraphics[width=0.32\textwidth]{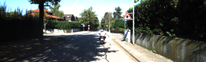} &
        \includegraphics[width=0.32\textwidth]{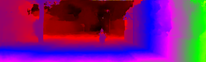} &
        \includegraphics[width=0.32\textwidth]{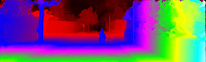} 
        \\
        
        \includegraphics[width=0.32\textwidth]{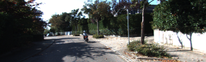} &
        \includegraphics[width=0.32\textwidth]{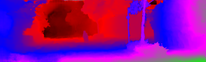} &
        \includegraphics[width=0.32\textwidth]{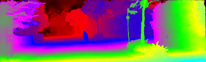} 
        \\
        
        \includegraphics[width=0.32\textwidth]{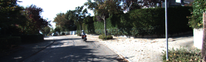} &
        \includegraphics[width=0.32\textwidth]{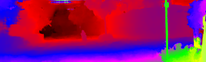} &
        \includegraphics[width=0.32\textwidth]{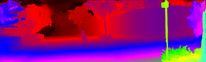} 
        \\
        
        \includegraphics[width=0.32\textwidth]{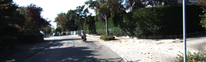} &
        \includegraphics[width=0.32\textwidth]{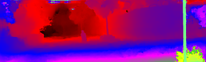} &
        \includegraphics[width=0.32\textwidth]{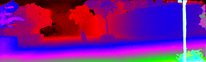} 
        \\
        
        \includegraphics[width=0.32\textwidth]{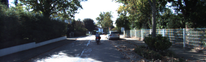} &
        \includegraphics[width=0.32\textwidth]{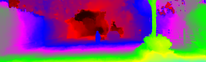} &
        \includegraphics[width=0.32\textwidth]{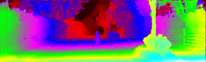} 
        \\
        (a) & (b) & (c)\\
    \end{tabular}
    \vspace{-8pt}
    \caption{Qualitative results on KITTI raw sequence \textit{2011\_10\_03\_drive\_0034\_sync}, for networks trained on \textit{sf-all}. (a) Left input frame. (b) Disparity map estimated by GCNet \cite{kendall2017-gcnet}. (c) Disparity map estimated by our \family-GCNet. Both networks are trained on the same synthetic data \textit{sf-all}.}
    \label{fig:supp-kt-raw-msgcnet}
      \vspace{-8pt}
\end{figure*}
\fi

\iftrue
\begin{figure*}[hbt!]
    \centering
    \renewcommand{\tabcolsep}{1pt}
    \scriptsize
    \begin{tabular}{ccc}
        \includegraphics[width=0.32\textwidth]{imgs/qualitatives/0000000101_L.png} &
        \includegraphics[width=0.32\textwidth]{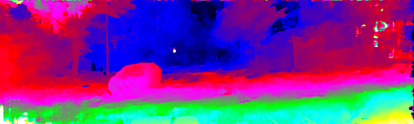} &
        \includegraphics[width=0.32\textwidth]{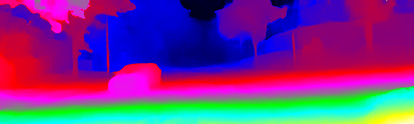} \\
        
        \includegraphics[width=0.32\textwidth]{imgs/qualitatives/0000000111_L.png} &
        \includegraphics[width=0.32\textwidth]{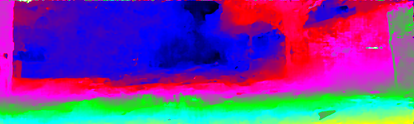} &
        \includegraphics[width=0.32\textwidth]{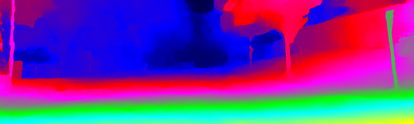} 

        \\
        
        \includegraphics[width=0.32\textwidth]{imgs/qualitatives/0000000195_L.png} &
        \includegraphics[width=0.32\textwidth]{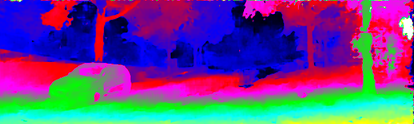} &
        \includegraphics[width=0.32\textwidth]{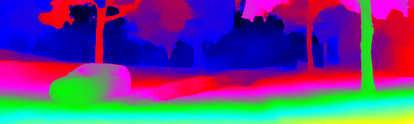} 
        
        \\
        
        \includegraphics[width=0.32\textwidth]{imgs/qualitatives/0000000381_L.png} &
        \includegraphics[width=0.32\textwidth]{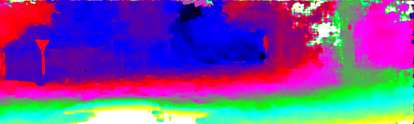} &
        \includegraphics[width=0.32\textwidth]{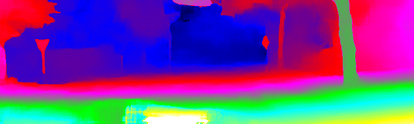} 
  \\
        
        \includegraphics[width=0.32\textwidth]{imgs/qualitatives/0000000991_L.png} &
        \includegraphics[width=0.32\textwidth]{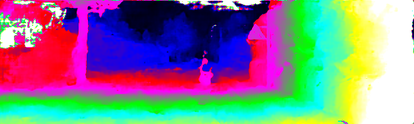} &
        \includegraphics[width=0.32\textwidth]{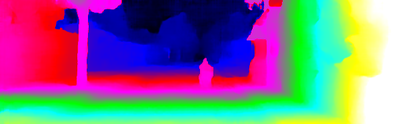} 
        \\
        
        \includegraphics[width=0.32\textwidth]{imgs/qualitatives/0000001057_L.png} &
        \includegraphics[width=0.32\textwidth]{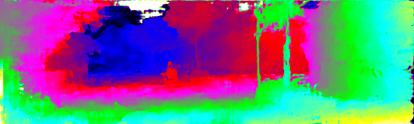} &
        \includegraphics[width=0.32\textwidth]{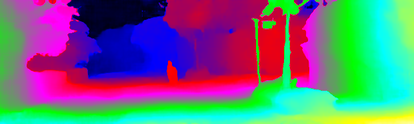} 
        \\
        
        \includegraphics[width=0.32\textwidth]{imgs/qualitatives/0000001062_L.png} &
        \includegraphics[width=0.32\textwidth]{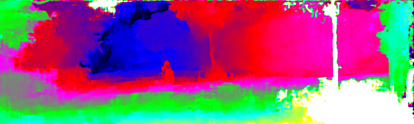} &
        \includegraphics[width=0.32\textwidth]{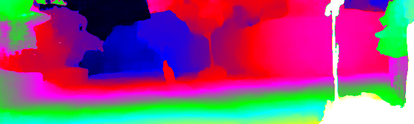} 
        \\
        
        \includegraphics[width=0.32\textwidth]{imgs/qualitatives/0000001063_L.png} &
        \includegraphics[width=0.32\textwidth]{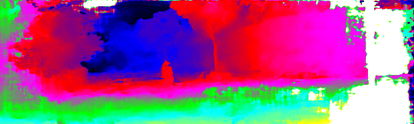} &
        \includegraphics[width=0.32\textwidth]{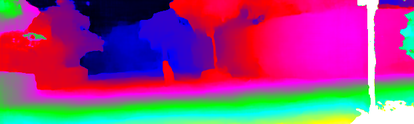} 
        \\
        
        \includegraphics[width=0.32\textwidth]{imgs/qualitatives/0000001118_L.png} &
        \includegraphics[width=0.32\textwidth]{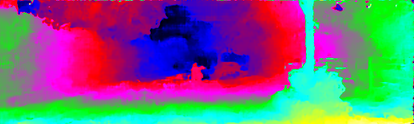} &
        \includegraphics[width=0.32\textwidth]{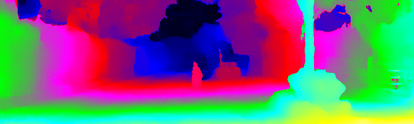} 
        \\
        (a) & (b) & (c)\\
    \end{tabular}
    \vspace{-8pt}
    \caption{Qualitative results on KITTI raw sequence \textit{2011\_10\_03\_drive\_0034\_sync}, for networks trained on \textit{sf-all}. (a) Left input frame. (b) Disparity map estimated by PSMNet \cite{chang2018psmnet}. (c) Disparity map estimated by our \family-PSMNet. Both networks are trained on the same synthetic data \textit{sf-all}.}
    \label{fig:supp-kt-raw-mspsmnet}
      \vspace{-8pt}
\end{figure*}
\fi

\end{document}